\documentclass[twocolumn, 10pt]{article}
\usepackage
[
        a4paper,
        left=1.5cm,
        right=1cm,
        top=3cm,
        bottom=3cm,
]
{geometry}

\usepackage{cite}
\usepackage[symbol]{footmisc}
\renewcommand{\thefootnote}{\fnsymbol{footnote}}
\usepackage{amsmath}
\usepackage{amsfonts}
\usepackage{amsthm}
\usepackage{physics}
\newtheorem{theorem}{Theorem}

\usepackage{array}
\usepackage[caption=false,font=footnotesize]{subfig}

\usepackage{url}

\usepackage[labelfont=bf]{caption} 
\usepackage[toc,page]{appendix}
\usepackage[affil-it]{authblk}
\usepackage{hyperref}
\hypersetup{colorlinks,urlcolor=blue}
\usepackage{multirow}
\usepackage{hhline}
\usepackage{booktabs}
\usepackage{xcolor}
\usepackage{contour}
\usepackage[normalem]{ulem}
\usepackage{algorithm}
\usepackage{algpseudocode}

\contourlength{0.8pt}
\renewcommand{\underline}[1]{%
  \uline{\phantom{#1}}%
  \llap{\contour{white}{#1}}%
}

\usepackage{dcolumn}
\newcolumntype{d}[1]{D{.}{.}{#1}}

\definecolor{MyRed}{HTML}{9C3300}
\definecolor{MyGreen}{HTML}{009C33}
\definecolor{MyPurple}{HTML}{33009C}

\DeclareMathOperator{\supp}{supp}
\DeclareMathOperator{\KL}{KL}

\usepackage{tikz}
\usetikzlibrary{decorations,backgrounds}

\def\pgfdecoratedcontourdistance{0pt}

\pgfkeys{/pgf/decoration/contour distance/.code={%
    \pgfmathparse{#1}%
    \let\pgfdecoratedcontourdistance=\pgfmathresult}%
}

\pgfdeclaredecoration{contour lineto}{start}
{
    \state{start}[next state=draw, width=0pt]{
        \pgfpathmoveto{\pgfpoint{0pt}{\pgfdecoratedcontourdistance}}%
    }
    \state{draw}[next state=draw, width=\pgfdecoratedinputsegmentlength]{
        \pgfmathparse{-\pgfdecoratedcontourdistance*cot(-\pgfdecoratedangletonextinputsegment/2+90)}%
        \let\shorten=\pgfmathresult%
        \pgfpathlineto{\pgfpoint{\pgfdecoratedinputsegmentlength+\shorten}{\pgfdecoratedcontourdistance}}%
    }
    \state{final}{
        \pgfpathlineto{\pgfpoint{\pgfdecoratedinputsegmentlength}{\pgfdecoratedcontourdistance}}%
    }
}

\newcommand{\algrule}[1][.2pt]{\par\vskip.5\baselineskip\hrule height #1\par\vskip.5\baselineskip}

\newcommand\blfootnote[1]{%
  \begingroup
  \renewcommand\thefootnote{}\footnote{#1}%
  \addtocounter{footnote}{-1}%
  \endgroup
}

\title{\huge \textbf{Diffeomorphic Counterfactuals\\with Generative Models}}

\author[1]{Ann-Kathrin~Dombrowski$^*$}
\author[1]{Jan~E.~Gerken$^*$}
\author[1234]{Klaus-Robert~M\"uller$^{\dagger}$}
\author[1]{Pan~Kessel$^{\dagger}$}
\affil[1]{Berlin Institute of Technology (TU
Berlin), 10587 Berlin, Germany}
\affil[2]{Department of Artificial Intelligence,
Korea University, Seoul 136-713, Korea}
\affil[3]{Max Planck Institut f\"ur Informatik, 66123 Saarbr\"ucken, Germany}
\affil[4]{Google Research, Brain Team}
\date{}
\begin{document}

\maketitle

\begin{abstract}
    Counterfactuals can explain classification decisions of neural networks in a human interpretable way. We propose a simple but effective method to generate such counterfactuals. More specifically, we perform a suitable diffeomorphic coordinate transformation and then perform gradient ascent in these coordinates to find counterfactuals which are classified with great confidence as a specified target class. We propose two methods to leverage generative models to construct such suitable coordinate systems that are either exactly or approximately diffeomorphic. We analyze the generation process theoretically using Riemannian differential geometry and validate the quality of the generated counterfactuals using various qualitative and quantitative measures.\blfootnote{\textbf{under review}}
\end{abstract}
\blfootnote{${}^*$ equal contribution}
\blfootnote{${}^{\dagger}$ corresponding authors: 
pan.kessel@tu-berlin.de, \newline
\hspace*{3.9cm} klaus-robert.mueller@tu-berlin.de}

\renewcommand*{\thefootnote}{\arabic{footnote}}

\section{Introduction}\label{sec:introduction}

Deep neural network models are widely used to solve complex problems from computer vision (e.g.~\cite{he2016resnet, ren2017faster, chen2017deeplab,lecun2015deep,schmidhuber2015deep}), strategic games and robotics (e.g.~\cite{silver2016mastering,thrun2005probabilistic,won2020adaptive}),  to medicine (e.g.~\cite{lengauer2007bioinformatics,capper2018dna,binder2021morphological}) and the sciences (e.g.\cite{baldi2014searching,noe2019boltzmann,unke2021,unke2021spookynet,degrave2022magnetic}). However, they are traditionally seen as black-box models, i.e.\ given  the network model, it has been unclear to the user and even the engineer designing the algorithm, what has been most important to reach a particular output prediction. This can cause serious obstacles for applications since, say, networks using spurious image features that are only present in the training data (Clever Hans effect~\cite{lapuschkin2019unmasking, anders2022clever}) might go unnoticed. Such undesired behaviour hampering the network's generalization ability is particularly problematic in safety-critical areas. 

Supplying this desired transparency has been the subject of recent developments in the field of explainable AI (XAI)~\cite{baehrens2010how,adadi2018peeking, samek2019explainable, samek2021explaining,holzinger2019causability}, ameliorating the aforementioned challenges. Prominent techniques in this area~\cite{baehrens2010how, simonyan2014deepInsideCNN, zeiler2014visualizing, springenberg2015gbp, bach2015pixelwise, ribeiro2016LIME, zintgraf2017visualizingDeepNNDecisions, shrikumar2017learningImportantFeatures, lundberg2017SHAP, dabkowski2017realTimeImageSaliency, smilkov2017smoothGrad, sundararajan2017axiomaticAttribution, fong2017interpretable, montavon2017deepTaylor, kindermans2018patternAttrNet, kim2017interpretability} construct e.g.\ saliency maps for classifiers or regressors \cite{letzgus2021toward} which highlight areas of the input that were particularly important for the classification.

\begin{figure}[t!]
\captionsetup[subfigure]{labelformat=empty, justification=centering}
\centering
\subfloat[original\\not blond]{\includegraphics[width=.3\linewidth]{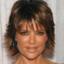}}
\hskip 0.2cm
\centering
\subfloat[adversarial example\\blond $(p\approx0.99)$]{\includegraphics[width=.3\linewidth]{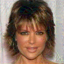}}
\hskip 0.2cm
\subfloat[counterfactual\\blond $(p\approx0.99)$]{\includegraphics[width=.3\linewidth]{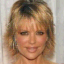}}
\caption{Example of a counterfactual from the CelebA dataset. The original is classified as not blond. The adversarial is classified with high confidence as blond, but the difference to the original resembles unstructured noise. The counterfactual is also classified with high confidence as blond but in contrast to the adversarial example it shows semantic differences to the original.
}
\label{fig:counterfactual_example}
\end{figure}

A different approach to explain a neural network is given by providing counterfactuals to the original inputs~\cite{miller2019explanation, verma2020counterfactual, stepin2021survey}. These are realistic-looking images which are semantically close to the original but differ in distinct features so that their classification matches the desired target class, cf.~Figure~\ref{fig:counterfactual_example}. Counterfactuals aim to answer questions like ``Why was this input classified as $A$ and not as $B$?'' or ``What would need to change in the input so that it is no longer classified as $A$ but instead as $B$?''~\cite{miller2019explanation, verma2020counterfactual, stepin2021survey} and thereby provide an explanation for the classifier. Unlike attribution methods, counterfactuals do not provide a relevance map, but an image that is similar to the original input and serves as a kind of counter example or hypothetical alternative for the original prediction.

\begin{figure*}[htp!]
\centering
\includegraphics[width=1.0\linewidth]{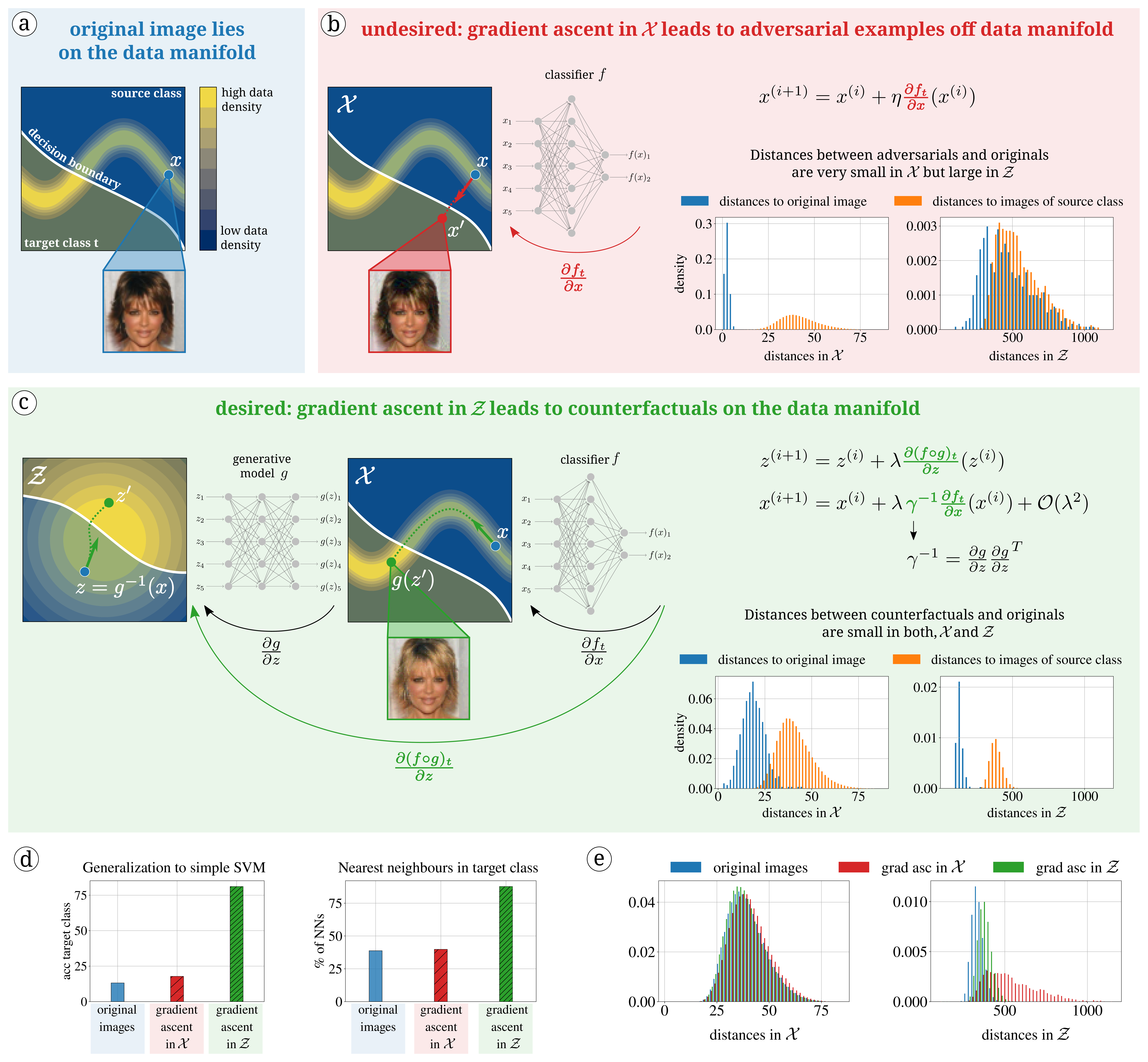}
\caption{\textbf{(a)} Image data usually lies on a lower dimensional data manifold, which is embedded in high dimensional space. We want to know what image features would have to change so that the classification flips. \textbf{(b)} If we follow the gradient of our target class with respect to the input $\frac{\partial f_t}{\partial x}$ the prediction flips but the resulting image is an adversarial example that looks indistinguishable from the original for a human observer. The changes to the original image are not semantic, but are limited to specific noise. The Euclidean difference between adversarial and original is therefore very small when measured in $\mathcal{X}$ but large when measured in $\mathcal{Z}$. \textbf{(c)} We use to the normalizing flow $g$ to obtain the latent space representation $z = g^{-1}(x)$ of our original image $x$. We then perform gradient ascent in the latent space $\mathcal{Z}$. The prediction flips, but this time the resulting image is a counterfactual. The changes to the original image are semantic. The Euclidean difference between counterfactual and original is small when measured in $\mathcal{X}$ and $\mathcal{Z}$. \textbf{(d)} Left: Quantitative evaluations show that counterfactuals generalize to simple classifiers in contrast to adversarial examples. Right: Counterfactuals are similar to images of the target class. We show results for the CelebA data set. \textbf{(e)} Histograms for original images, adversarial examples and counterfactuals are indistinguishable when measuring the distances in the input space $\mathcal{X}$. When measuring the distances in the latent space $\mathcal{Z}$ we see that adversarial examples have larger distances. This confirms the hypothesis that adversarial examples lie off manifold. We show results for the CelebA data set.}
\label{fig:overview}
\end{figure*}

Crucially, the counterfactual is required to be a realistic sample from the data distribution in order to elucidate the behaviour of the network on the data. This requirement poses the greatest practical challenge to computing counterfactuals since naively optimizing the output of the network with respect to the input via gradient ascent yields adversarial examples~\cite{szegedy2013intriguing} which essentially add a small amount of noise to the original input, as illustrated in the example given in Figure~\ref{fig:counterfactual_example}. This behavior can be understood using the manifold hypothesis: the images are assumed to lie on a low dimensional manifold embedded in the high dimensional input space, cf.~Figure~\ref{fig:overview}~(a). The gradient ascent algorithm then walks in a direction orthogonal to the decision boundary which is with high probability also orthogonal to the data manifold, resulting in a small perturbation which is not semantic, as illustrated in Figure~\ref{fig:overview}~(b).

We propose to use insights from the mathematical discipline of differential geometry to mitigate this problem. Differential geometry can be understood as analysis on curved (hyper-) surfaces and thus provides the appropriate tools to study gradients on the data manifold. It has been valuable for the field of ML in general~\cite{amari2016information, bronstein2017geometric, shao2018riemannian} and XAI specifically~\cite{dombrowski2019explanations,anders2020fairwashing, dombrowski2022towards}. 
 A cornerstone of differential geometry is the idea that geometric quantities can be described equivalently in different coordinate systems. However, not all coordinate systems are equally useful in practice. This phenomenon is ubiquitous in physics: for instance, the mathematical expressions governing planetary motions greatly simplify in a heliocentric (sun-centered) coordinate system as opposed to a geocentric (earth-centered) coordinate system. In heliocentric coordinates, the relevant degrees of freedom are easily recognizable (Keppler ellipses) and, as a result, physical intuition and interpretations are much easier to deduce. Similarly, we attribute the difficulty to construct counterfactuals by optimizing the output of a neural network classifier with respect to its input as in Figure~\ref{fig:overview}~(b) to the poor choice of coordinates $\mathcal{X}$ in the input space given by the raw image data. In contrast, in a suitably chosen coordinate system $\mathcal{Z}$, the data manifold would extend more evenly in all directions, allowing for an optimization that stays on the data manifold and thereby producing a counterfactual that has been changed {\em semantically} when compared to the original. In order to find such a coordinate transformation (called a \emph{diffeomorphism} in differential geometry) between $\mathcal{X}$ and $\mathcal{Z}$, we use a normalizing flow trained on the image data set under consideration. Since the flow is by construction bijective and differentiable with a differentiable inverse, it satisfies the technical conditions for a diffeomorphism in differential geometry. Furthermore, the base distribution of the flow is fixed to be a univariate Gaussian and hence free of pathological directions. Moreover, this change of coordinate system will lead to no information loss which is in stark contrast to existing methods for generating counterfactuals.

In our method, the counterfactual is computed by taking the gradient in the gradient ascent update with respect to the representation in the base space of the normalizing flow as opposed to the input of the classifier (Figure~\ref{fig:overview}~(b)). This method comes with rigorous theoretical guarantees and we refer to it as {\em diffeomorphic counterfactuals}. In particular, we show that this introduces a metric into the update step which shrinks the gradient in directions orthogonal to the data manifold. Furthermore, we propose two separate methods which only {\em approximately} lead to a diffeomorphism. While these approximate methods come with a lower level of theoretical guarantees, and can, in practice, lead to some information loss, they can be scaled easily to very high-dimensional datasets, as we demonstrate experimentally. We refer to these methods as {\em approximate diffeomorphic counterfactuals}. We theoretically prove that these methods also stay on the data manifold under suitable assumptions. Our theoretical analysis therefore provides a unified mathematical framework for the application of generative models in the context of counterfactuals. Importantly, we can not only optimize the output of a classifier network on the data manifold in this manner, but also that of a regressor.

This analysis is supported by our experimental results for various application domains, such as computer vision and medical radiology and a number of architectures for classifiers, regressors and generative models. Note that we lay emphasis on using quantitative metrics --- as opposed to only qualitative analysis --- to evaluate the proposed methods; some quantitative results are exemplified in Figure~\ref{fig:overview}~(d) and (e).

The paper is structured as follows: in Section~\ref{sec:method}, we introduce the proposed methods. Specifically, we will introduce diffomorphic explanations in Section~\ref{sec:diffeoCounter} and the approximate versions thereof in Section~\ref{sec:approxDiffeo}. We then analyse the proposed methods theoretically using Riemannian differential geometry in Section~\ref{sec:theory}. This is followed by Section~\ref{sec:experiments} which provides a detailed experimental analysis of the diffeomorphic counterfactuals, in Section~\ref{sec:experiments_diffeo}, and approximate diffeomorphic counterfactuals in Section~\ref{sec:experiments_approx}. In Section~\ref{sec:relatedworks}, we give an extensive discussion of related work. The code for a toy example and our main experiments is publicly accessible\footnote{\url{https://github.com/annahdo/counterfactuals}}.

\section{Methods}\label{sec:method}
In this section, we will introduce in detail our novel diffeomorphic and approximately diffeomorphic counterfactuals. For this, we will start by reviewing the  basics of counterfactual explanations and then present our two proposed methods.

\subsection{Counterfactual Explanations}
Consider a classifier $f:\mathcal{X} \to \mathbb{R}^C$ which assigns to an input $x \in \mathcal{X}$ the probability $f(x)_c$ to be part of class $c \in \{1,\dots, C\}$. Counterfactual explanations of the classifier $f$ provide minimal deformations $x'=x+\delta x$ such that the prediction of the classifier is changed. 

In many cases of practical relevance, the data lies approximately on a submanifold $\mathcal{D}\subset\mathcal{X}$ which is of significantly lower dimensionality $N_\mathcal{D}$ than the dimensionality $N_\mathcal{X}$ of the input space $\mathcal{X}$. This is known as the manifold hypothesis in the literature (see e.g. \cite{GoodfellowBook}). For counterfactual explanations, as opposed to adversarial examples, we are interested in deformations $x'$ which lie on the data manifold. Additionally, we require the deformations to the original data to be minimal, i.e.\ the perturbation $\delta x$ should be as small as possible. The relevant norm is however measured along the data manifold and not calculated in the input space. For example, a slightly rotated number in an MNIST image may have large pixel-wise distance but should be considered an infinitesimal perturbation of the original image. 

We mathematically formalize the manifold hypothesis by assuming that the data is concentrated in a small region of extension $\delta$ around $\mathcal{D}$. As we will show in Section~\ref{sec:theory}, this implies that the support $S$ of the data density $p$ is a product manifold
\begin{align}
    S = \mathcal{D} \times \mathcal{I}_{\delta_1} \times \dots \times \mathcal{I}_{\delta_{N_\mathcal{X}-N_\mathcal{D}}} \,,\label{eq:prod_mfd}
\end{align}
where $\mathcal{I}_\delta=(-\frac{\delta}{2}, \frac{\delta}{2})$ is an open interval of length $\delta$ (with respect to the Euclidean distance on the input space $\mathcal{X}$). We assume that  $\delta$ is small, i.e. the data lies approximately on the low-dimensional manifold $\mathcal{D}$ and thus fulfills the manifold hypothesis. 
We can think of the $\mathcal{I}_\delta$ as arising from the inherent noise in the data. 

Furthermore, we define the set of points in $S$ classified with confidence $\Lambda \in (0, 1)$ as class $t \in \{1, \dots, C\}$ by
\begin{align}
    S_{t, \Lambda} = \{ x\in S \; | \; t = \textrm{argmax}_j f_j(x) \; \textrm{and} \; f_t(x) > \Lambda \}\,.
\end{align}
A \emph{counterfactual} $x'\in\mathcal{X}$ for class $t$ of the original sample $x\in\mathcal{X}$ then is the closest point to $x$ in $S_{t,\Lambda}$,
\begin{align}
    x'\in S_{t,\Lambda}\quad\text{and}\quad \textrm{argmin}_y d_\gamma(x,y)=x'\,,
\end{align}
where $d_\gamma(x', x)$ is the distance computed by the Riemannian metric $\gamma$ on $S$ (which is induced from the flat metric by the diffeomorphism given by the generative model). We will review the necessary concepts of Riemannian geometry in Section~\ref{sec:differentialgeometry}.

\subsection{Generation of Counterfactuals}
Often, counterfactuals are generated by performing gradient ascent in the input space $\mathcal{X}$, see \cite{verma2020counterfactual} for a recent review on counterfactuals. More precisely, for step size $\eta$ and target class $t$, one performs the gradient ascent step
\begin{align}
  x^{(i+1)}=x^{(i)}+\eta \frac{\partial f_{t}}{\partial x}(x^{(i)})\label{eq:2} 
\end{align}
until the classifier has reached a threshold confidence $\Lambda$, i.e. $f(x^{(i+1)})_t > \Lambda$. The resulting samples will however often not lie on the data manifold and differ from the original image $x$ only in added unstructured noise rather than in an interpretable and semantically meaningful manner. Especially when applied to high dimensional image data such samples are usually referred to as adversarial examples and not counterfactuals. 

We therefore propose to estimate the counterfactual $x'$ of the original data point $x$ by using a diffeomorphism $g: \mathcal{Z} \to S$. We then perform gradient ascent in the latent space $\mathcal{Z}$, i.e.
\begin{align}
      z^{(i+1)}=z^{(i)}+\lambda \frac{\partial(f\circ g)_{t}}{\partial z}(z^{(i)})\label{eq:1}
    \end{align}
with step size $\lambda \in \mathbb{R}_+$. This has the important advantage that the resulting counterfactual will lie on the data manifold. Furthermore, since we consider a diffeomorphism $g$, and thus in particular a bijective map, no information will be lost by considering the classifier $f \circ g$ on $\mathcal{Z}$ instead of the original classifier $f$ on the data manifold $S$, i.e. there exists a unique $z = g^{-1}(x) \in \mathcal{Z}$ for any $x \in S$. 
We show pseudo code for our approach in Algorithm~\ref{alg:generating_counterfactuals}.

\begin{figure}[t]
\centering
\includegraphics[width=1.0\linewidth]{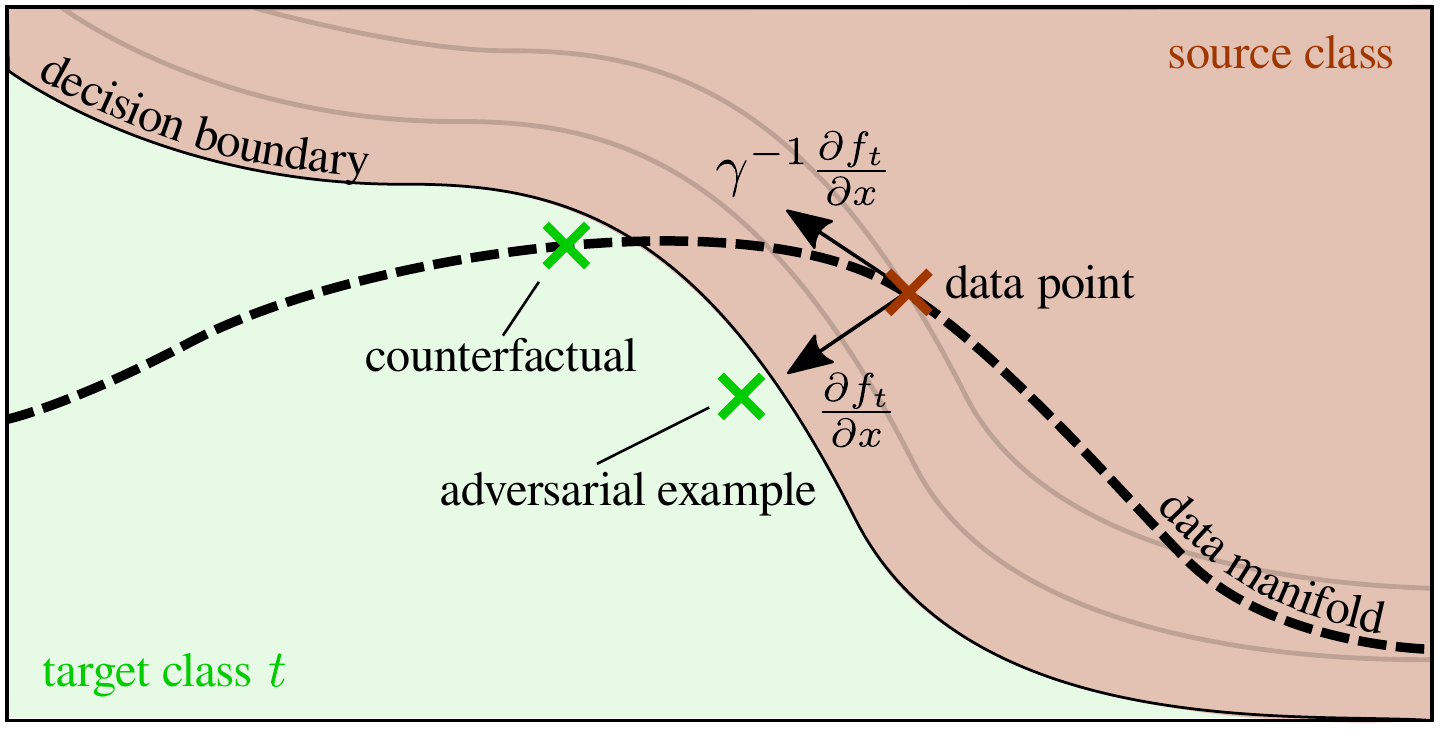}
\caption{When the gradient ascent optimization of the target class is performed in the input space of the classifier, one leaves the data manifold and obtains an adversarial example. If instead the gradient ascent is performed in the latent space of a generative model, one stays on the data manifold, resulting in a counterfactual example.}
\label{fig:method_intuition}
\end{figure}

As illustrated in Figure~\ref{fig:method_intuition}, gradient ascent in $\mathcal{X}$ and $\mathcal{Z}$ are well-suited to generate adversarial examples and counterfactuals, respectively.

\begin{algorithm}
\caption{Generating counterfactuals}\label{alg:generating_counterfactuals}
\begin{algorithmic}[1]
  \Require $x, f, g, g^{-1}, t, \Lambda, \lambda, N$
  \State $z \gets g^{-1}(x)$
  \For{$i$ in range($N$)}
    \State $\nabla_z \gets \frac{\partial (f\circ g)_t}{\partial z}$
    \State $z \gets $ optimizer.step($\lambda$, $\nabla_z$)
    \If{$f(g(z))_t>\Lambda$}
        \State \Return $g(z)$
    \EndIf
  \EndFor
  \State\Return None
\end{algorithmic}
\algrule[.5pt]
\noindent \textbf{Note:} $x$ is the input for which we desire to find a counterfactual explanation, $f$ the predictive model, $g$ the generative model, $g^{-1}$ the (approximate) inverse of $g$, $t$ the target class, $\Lambda$ the target confidence, $\lambda$ the learning rate and $N$ the maximum number of update steps. \vskip 2pt
\end{algorithm}

For regression tasks there is no explicit decision boundary, but we can still follow the the same algorithm by directly maximizing (or minimizing) the output $r$ of regressor $f(x)$ until we reach the desired target regression value.

\subsection{Novel Method 1: Diffeomorphic Counterfactuals}\label{sec:diffeoCounter}
We propose to model the map $g$ by a normalizing flow and will refer to the corresponding modified data $x'$ as diffeomorphic counterfactuals in the following. 

Specifically, a flow $g$ is an invertible neural network which equips, by the change-of-variable theorem, the input space $\mathcal{X}$ with a probability density 
\begin{align}
    q(x) = q_Z(g^{-1}(x)) \, \left|\textrm{det}\frac{\partial z}{\partial x}\right| \,,
\end{align}
where $q_Z$ is a simple base density, such as a univariate normal density, on the latent space $\mathcal{Z}$. The flow can be trained by maximum likelihood, i.e. by minimizing
\begin{align}
    \textrm{KL}(p | q) &= -\mathbb{E}_{x \sim p} \log q(x) + \textrm{const.}\nonumber\\
    &\approx - \frac{1}{N} \sum_{i=1}^N \log(q(x_i)) + \textrm{const.} \,,
\end{align}
where $x_i \sim p$ are samples from the data density $p$. Since the flow is bijective on the entire input space $\mathcal{X}$, it will, in particular, be bijective on the data manifold $S \subset \mathcal{X}$. Furthermore, we will also rigorously show in Section~\ref{sec:flows} that a well-trained flow maps (to very good approximation) only to the data manifold, i.e. $g(\mathcal{Z}) \approx S$. 

Therefore, flows guarantee that no information is lost when performing gradient ascent in the latent space $\mathcal{Z}$ and also ensure that the resulting counterfactuals lie on the data manifold $S$. Indeed, the flow can be understood as inducing a certain coordinate change of the input space $\mathcal{X}$ which is particularly suited for the generation of counterfactuals.

\subsection{Novel Method 2: Approximate Diffeomorphic Counterfactuals}\label{sec:approxDiffeo}
While the method of the last section is very appealing as it comes with strong guarantees, it may be challenging to scale to very high-dimensional data sets. This is because flows have a very large memory footprint on such datasets as each layer has the same dimensionality as the data space $\mathcal{X}$ to ensure bijectivity. 
We therefore posit an alternative method, called approximate diffeomorphic counterfactuals, which comes with less rigorous theoretical guarantees, but can scale better to very high-dimensional data. Specifically, we propose two varieties of approximate diffeomorphic counterfactuals:

\textbf{Autoencoder-based:} the reconstruction loss of an autoencoder (AE), i.e.
\begin{align}
    \mathcal{L} = \mathbb{E}_{x \sim p} ||g(e(x)) - x ||^2 \,, 
\end{align}
with encoder $e:\mathcal{X} \to \mathcal{Z}$ and generator $g:\mathcal{Z} \to \mathcal{X} $ is minimized if the encoder is the inverse of the generator on the data manifold $S$,
i.e.
\begin{align}
 e|_S = g^{-1}|_S   \,.
\end{align}
This implies, in particular, that $g(\mathcal{Z})=S$ if $\textrm{dim}(\mathcal{Z}) = \textrm{dim}(S)$.
As for normalizing flows, the image of the autoencoder is the data manifold if the model has been perfectly trained. However, an autoencoder will only be invertible on the data manifold in this perfect training limit and if the latent space $\mathcal{Z}$ has the same dimension as the data space $S$. This is in contrast to normalizing flows which are invertible on all of $\mathcal{X}$ by construction. As a result, the autoencoder will necessarily lead to loss of information unless the model is perfectly trained and the latent space dimensionality perfectly matches the dimension of the data.

\textbf{GAN-based:} Generative Adversarial Networks (GANs) consist of a generator $g:\mathcal{Z} \to \mathcal{X}$ and a discriminator $d:\mathcal{X} \to \{0,1\}$. Training then proceeds by minimizing a certain minimax loss, see \cite{goodfellow2014generative} for details. It can be shown that the global minimizer of this loss function ensures that samples of the optimal generator $g$ are distributed according to the data distribution, i.e.
\begin{align}
    g(z) \sim p \quad \textrm{for} \quad z \sim q_Z \,.
\end{align}
We refer to Section~4.1 of \cite{goodfellow2014generative} for a proof.
However, the optimal generator $g$ is not necessarily bijective on the data manifold. 
This implies that even for a perfectly trained GAN, there may not exist a unique $z \in \mathcal{Z}$ for a given data sample $x \in \mathcal{X}$ such that $x=g(z)$. Furthermore, there is no manifest mechanism to obtain the corresponding latent sample $z \in \mathcal{Z}$ for a given input $x \in \mathcal{X}$. This is in contrast to normalizing flows and autoencoders, since, for these generative models, the inverse map $g^{-1}:\mathcal{X} \to \mathcal{Z}$ is either explicitly or approximately known, respectively. 

However, there is an extensive literature for GAN inversion, see \cite{xia2021ganinversion} for a recent review. For a given generator $g$ and data sample $x \in \mathcal{X}$, these methods aim to find a latent vector $z \in \mathcal{Z}$ such that $x \approx g(z)$. This is often done by minimizing the difference between the activations of an intermediate layer of some auxiliary network, i.e.
\begin{align}
    z = \textrm{argmin}_{\hat{z} \in \mathcal{Z}}|| h(g(\hat{z})) - h(x) || \,.
\end{align}
For example, $h$ can be chosen to be an intermediate layer of an Inception network \cite{szegedy2015rethinking} trained on samples from the data density $p$. Note that  these inversion methods do not come with rigorous guarantees as the optimization objective is non-convex and it is unclear whether the values of the intermediate layer activations are sufficient to distinguish different inputs.

\section{Theoretical analysis}
\label{sec:theory}

In this section, we employ tools from differential geometry to show that for well-trained generative models, the gradient ascent update \eqref{eq:1} in the latent space $\mathcal{Z}$ does indeed stay on the data manifold, as confirmed by the experimental results presented in Section~\ref{sec:experiments}. Intuitively, since in \eqref{eq:1} we take small steps in $\mathcal{Z}$, where the probability distribution is, for example, a normal with unit variance, we do not leave the region of high probability in the latent space and hence stay in a region of high probability also in $\mathcal{X}$. 

We prove this statement for the case of diffeomorphic counterfactuals, i.e. for normalizing flows, and -- under stronger assumptions -- also for approximate diffeomorphic counterfactuals, i.e. for autoencoders and generative adversarial networks.

\subsection{Differential Geometry}\label{sec:differentialgeometry}
In this section, we briefly introduce the most fundamental notions of differential geometry used in our discussion further down. For a comprehensive textbook, see e.g.~\cite{lee2012}.

Differential geometry is the study of smooth (hyper-)surfaces. The central notion of this branch of mathematics is that of an $n$-dimensional (differentiable) manifold $\mathcal{M}$ which is equipped with coordinate functions $x^{\mu}:\mathcal{M}\rightarrow\mathbb{R}^{n}$ (so-called \emph{charts} which are assembled into an \emph{atlas}). These coordinates allow for explicit calculations in $\mathbb{R}^{n}$, but geometric objects (tensors) are independent of the chosen coordinates and transform in well-defined ways under changes of coordinates. Such a change of coordinates can be interpreted as a differentiable bijection $\phi:\mathcal{M}\rightarrow\mathcal{M}$ whose inverse is also differentiable, a so-called \emph{diffeomorphism}.

At each point $p\in\mathcal{M}$, we attach an $n$-dimensional vector space $T_{p}\mathcal{M}$, the \emph{tangent space} at $p$. 
Coordinates $x^{\mu}$ induce a basis in $T_{p}\mathcal{M}$ and we will denote the components of $v\in T_{p}\mathcal{M}$ in this basis by $v^{\mu}$. Under a diffeomorphism $\phi$, the components of $v$ transform as
\begin{align}
  v^{\alpha}=\sum_{\mu=1}^{n}\frac{\partial \phi^{\alpha}}{\partial x^{\mu}}v^{\mu}\,.
\end{align}
To capture the notion of distance (and curvature) on $\mathcal{M}$, a \emph{metric tensor} $\gamma(p):T_{p}\mathcal{M}\times T_{p}\mathcal{M}\rightarrow\mathbb{R}$ is used. The metric defines a canonical isomorphism between $T_{p}\mathcal{M}$ and its dual space $T^{*}_{p}\mathcal{M}$. Following the usual convention in the general relativity literature, we use lower indices $v_{\mu}$ to denote the components of the dual vector $\gamma(p)(v,\cdot)$. This implies that contraction with the metric is used to raise and lower indices,
\begin{align}
  v_{\mu}=\gamma_{\mu\nu}v^{\nu}\quad\text{and}\quad v^{\mu}=\gamma^{\mu\nu}v_{\nu}\,,
\end{align}
where we used the Einstein summation convention to sum over repeated upper and lower indices and introduced $\gamma^{\mu\nu}$ for the inverse of $\gamma_{\mu\nu}$, the components of $\gamma$ in the basis induced by the coordinates $x^{\mu}$.

Given a metric, it is natural to consider shortest paths between points on $\mathcal{M}$. The corresponding curves are called \emph{geodesics}. If the length of the tangent vector of a geodesic $\sigma$ is constant (as measured by the metric) along $\sigma$, the geodesic is \emph{affinely-parametrized}. Importantly, the notion of an affinely parametrized geodesic is coordinate independent and can therefore itself be used to construct coordinates on $\mathcal{M}$, as we will see below.

\subsection{Mathematical Setup}
\label{sec:math-setup}
In order to analyze the gradient ascent \eqref{eq:1} in the latent space $\mathcal{Z}$, we define in this section the necessary manifolds and coordinates.

As above, let $\mathcal{X}$ be an $N_{\mathcal{X}}$-dimensional manifold which is the input space of the classifier $f:\mathcal{X}\rightarrow\mathbb{R}^{C}$ with $C$ classes. An implementation of the classifier corresponds to a function on $\mathbb{R}^{N_{\mathcal{X}}}$ and we denote the coordinates on $\mathcal{X}$ in which our classifier is given by $x^{\alpha}$. These coordinates could e.g.\ be suitably normalized pixel values. We furthermore use an $N_{\mathcal{Z}}$-dimensional manifold $\mathcal{Z}$ as the latent space for our generative model $g:\mathcal{Z}\rightarrow\mathcal{X}$. For GANs and AEs, we typically have $N_{\mathcal{Z}} < N_{\mathcal{X}}$ and for normalizing flows $N_{\mathcal{Z}} = N_{\mathcal{X}}$. In the latter case we have moreover $\mathcal{X}=\mathcal{Z}$ and $g$ bijective with differentiable inverse implying that $g$ is a diffeomorphism. Similarly to the classifier, also the generative model is implemented in specific coordinates on $\mathcal{Z}$ which we denote by $z^{a}$.

We equip $\mathcal{Z}$ with a flat Euclidean metric $\delta_{ab}$. Then, the generative model $g$ induces an inverse metric $\gamma^{\alpha\beta}$ on $g(\mathcal{Z})$ by
\begin{align}
  \gamma^{\alpha\beta}=\delta^{ab}\frac{\partial g^{\alpha}}{\partial z^{a}}\frac{\partial g^{\beta}}{\partial z^{b}}\,.\label{eq:invmetric}
\end{align}
in the case of $N_{\mathcal{Z}}<N_{\mathcal{X}}$, $\gamma$ is singular. This metric is the crucial new ingredient when performing the gradient ascent update in the latent space \eqref{eq:1} as opposed to in the input space \eqref{eq:2}, as the following calculation shows.

One step of gradient ascent in the latent space $\mathcal{Z}$ is given by the image under $g$ of the update step \eqref{eq:1}. In $x^{\alpha}$ coordinates and to linear order in the learning rate $\lambda$, it is given by
\begin{align}
  g^{\alpha}(z^{(i+1)})&=g^{\alpha}(z^{(i)})+\lambda\,\frac{\partial g^{\alpha}}{\partial z^{a}}\frac{\partial(f\circ g)_{t}}{\partial z^{a}}+\mathcal{O}(\lambda^{2})\nonumber\\
  &=g^{\alpha}(z^{(i)})+\lambda\,\frac{\partial g^{\alpha}}{\partial z^{a}}\frac{\partial g^{\beta}}{\partial z^{a}}\frac{\partial f_{t}}{\partial x^{\beta}}+\mathcal{O}(\lambda^{2})\nonumber\\
  &=g^{\alpha}(z^{(i)})+\lambda\,\gamma^{\alpha\beta}\frac{\partial f_{t}}{\partial x^{\beta}}+\mathcal{O}(\lambda^{2})\,.\label{eq:4}
\end{align}
If we start from the same points, $x^{(i)}=g(z^{(i)})$, the difference between gradient ascent in latent space \eqref{eq:1} and input space \eqref{eq:2} is just given by the contraction of the gradient of $f$ with respect to $x$ with the inverse induced metric $\gamma^{\alpha\beta}=\tfrac{\partial g^{\alpha}}{\partial z^{a}}\tfrac{\partial g^{\beta}}{\partial z^{a}}$. Hence, in order to understand why the prescription \eqref{eq:1} stays on the data manifold, we will in the following investigate the properties of $\gamma$ for the case of well-trained generative models.

Before returning to $\gamma$, we will first discuss the structure of the data. The probability density of the data on $\mathcal{X}$ is denoted by $p:\mathcal{X}\rightarrow\mathbb{R}$ and the probability density induced by $g$ is denoted by $q:\mathcal{X}\rightarrow\mathbb{R}$. For $q$ in $x^{\alpha}$ coordinates, we use the notation $q_{x}:\mathbb{R}^{N_{\mathcal{X}}}\rightarrow\mathbb{R}$. The data is characterized by $S=\supp(p)\subset\mathcal{X}$ which becomes $S_{x}\subset\mathbb{R}^{N_{\mathcal{X}}}$ in $x^{\alpha}$ coordinates. We will assume that the data lives approximately on a submanifold $\mathcal{D}\subset S$ of $\mathcal{X}$ with dimension $N_{\mathcal{D}}\ll N_{\mathcal{X}}$. In relation to the dimension of our generative model, we assume that $N_{\mathcal{D}}\leq N_{\mathcal{Z}}\leq N_{\mathcal{X}}$. As a subset of $\mathcal{X}$ and in $x^{\alpha}$ coordinates, $\mathcal{D}$ will be denoted by $\mathcal{D}_{x}\subset\mathbb{R}^{N_{\mathcal{X}}}$. To capture that the data does not extend far beyond $\mathcal{D}$, we assume that $S$ has Euclidean extension $\delta\ll 1$, normal to $\mathcal{D}$ in $x^{\alpha}$ coordinates, i.e.\footnote{The form \eqref{eq:5} restricts the slices $S_\perp(x_\mathcal{D})$ through $S$ normal to $\mathcal{D}$ to be L\textsubscript{1} balls whose size is independent of $x_\mathcal{D}$. We make this restriction to simplify notation but the argument can straightforwardly be extended to arbitrary shapes of $S_\perp(x_\mathcal{D})$ by bounding it by an L\textsubscript{2} ball of radius $\delta/2$.}
\begin{align}
  S_{x}=\left\{ x_{\mathcal{D}}+x_{\delta}\ \Big|\ x_{\mathcal{D}}\in \mathcal{D}_{x},\ x_{\delta}^{\alpha}\in\Big(-\frac{\delta}{2},\frac{\delta}{2}\Big) \right\}\,.\label{eq:5}
\end{align}

\begin{figure}[t]
  \centering
  \begin{tikzpicture}[scale=2.5,show background rectangle, line width=0.8pt]
    \draw[postaction={decoration={contour lineto, contour distance=-15},draw=MyGreen, dashed, decorate},
          postaction={decoration={contour lineto, contour distance=15},draw=MyGreen, dashed, decorate},
          domain=2.1416:4.1416,samples=100,MyGreen] plot (\x, {0.5*tan(-\x r)});
    \node at (4.23,-0.98) [MyGreen,anchor=base] {$S$};
    \node[MyGreen,rotate=-58,scale=1] at (4.16,-0.82) {$\left.\vphantom{\raisebox{1.5em}{x}}\right\}$};
    \node[black] at (4.2916,0.77) {$\mathcal{X}$};
    \node[MyGreen,anchor=west] at (3.9,-0.4) {$\mathcal{D}$};
    \draw[black,domain=2.6416:3.6416] plot (\x, {0.5*tan(-\x r)});
    \node at (2.5616,0.2) [black,anchor=base] {$y_{\parallel}$};
    \begin{scope}
      \clip (2.7916,0.55) -- ++(1,-0.5) -- ++(-0.3,-0.6) -- ++(-1,0.5) -- ++(0.3,0.6);
      \begin{scope}[shift={(3.1416,0)}]
        \draw[gray] (-0.5,-0.5) .. controls (-0.2,-0.4) and (-0.1,-0.2) ..
                    (0,0) .. controls (0.1,0.2) and (0.2,0.4) .. (0.5,0.5);
      \end{scope}
      \begin{scope}[shift={(3.3416,-0.02)}]
        \draw[gray] (-0.5,-0.5) .. controls (-0.2,-0.4) and (-0.1,-0.2) ..
                    (0,0) .. controls (0.1,0.2) and (0.2,0.4) .. (0.5,0.5);
      \end{scope}
      \begin{scope}[shift={(2.9416,0.02)}]
        \draw[gray] (-0.5,-0.5) .. controls (-0.2,-0.4) and (-0.1,-0.2) ..
                    (0,0) .. controls (0.1,0.2) and (0.2,0.4) .. (0.5,0.5);
      \end{scope}
    \end{scope}
    \begin{scope}[shift={(3.1416,0)}]
      \draw[gray,dotted] (-0.5,-0.5) .. controls (-0.2,-0.4) and (-0.1,-0.2) ..
                  (0,0) .. controls (0.1,0.2) and (0.2,0.4) .. (0.5,0.5);
    \end{scope}
    \begin{scope}[shift={(3.3416,-0.02)}]
      \draw[gray,dotted] (-0.5,-0.5) .. controls (-0.2,-0.4) and (-0.1,-0.2) ..
                  (0,0) .. controls (0.1,0.2) and (0.2,0.4) .. (0.5,0.5);
      \node at (0.7,0.4) [gray,anchor=base] {$y_{\bot}$};
    \end{scope}
    \begin{scope}[shift={(2.9416,0.02)}]
      \draw[gray,dotted] (-0.5,-0.5) .. controls (-0.2,-0.4) and (-0.1,-0.2) ..
                  (0,0) .. controls (0.1,0.2) and (0.2,0.4) .. (0.5,0.5);
    \end{scope}
    \begin{scope}[shift={(3.1416,0)}]
      \draw[MyPurple] (0,0) -- +(-0.0975,-0.195) node[anchor=west] {$\frac{\delta}{2}$};
    \end{scope}
    \begin{scope}[shift={(3.1416,0)}]
      \draw[MyRed] (0,0) -- node[anchor=east,xshift=2,yshift=2]{$\sigma$} (0.0805,0.155);
      \draw[fill] (0,0) circle (0.02) node[anchor=north east,yshift=4]{$p$};
      \draw[fill] (0.0775,0.155) circle (0.02) node[anchor=west]{$q$};
    \end{scope}
  \end{tikzpicture}
  \caption{Construction of the $y^{\mu}$ coordinates which are aligned with the data manifold $\mathcal{D}$.}
  \label{fig:y_coords}
\end{figure}
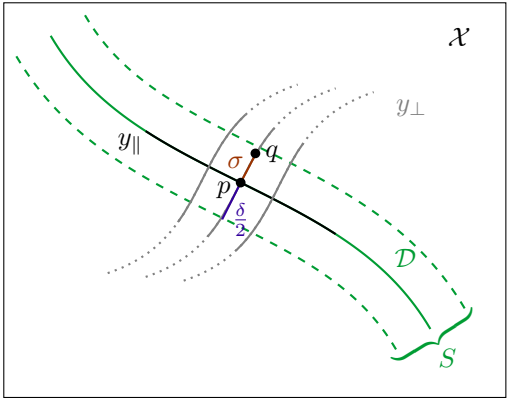

Next, we will define coordinates in a neighborhood of $\mathcal{D}$ which separate the directions tangential and normal to $\mathcal{D}$ as illustrated in Figure~\ref{fig:y_coords}. Our construction is similar to the constructions of Riemannian and Gaussian normal coordinates, adapted for a submanifold of codimension larger than one. First, we choose coordinates $y_{\parallel}$ on $\mathcal{D}$ and, for each $p\in\mathcal{D}$, a basis $\{n_{i}\}$ for the tangent space $T_{p}\mathcal{D}_{\perp}$ of the normal to $\mathcal{D}$ at $p$. Following the usual construction of Riemannian normal coordinates, we assign coordinates to a point $q$ in some neighborhood of $\mathcal{D}$ by constructing an affinely parametrized geodesic $\sigma:[0,1]\rightarrow \mathcal{X}$ which satisfies $\sigma(0)=p$ and $\sigma(1)=q$ and which has tangent vector $\sigma'(0)\in T_{p}\mathcal{D}_{\perp}$. The coordinates of $q$ are then $y(q)=(y_{\parallel}(p),y_{\perp})\in \mathbb{R}^{N_{\mathcal{D}}} \oplus \mathbb{R}^{N_{\mathcal{X}}-N_{\mathcal{D}}}$, where the $i$\textsuperscript{th} component of $y_{\perp}$ is given by the $i$\textsuperscript{th} component of $\sigma'(0)$ in the basis $\{n_{i}\}$. In a sufficiently small neighborhood around $\mathcal{D}$, we can find a unique basepoint $p\in\mathcal{D}$ and geodesic $\sigma$ for every $q$.

One important aspect of this construction is that by rescaling the basis vectors $\{n_{i}\}$, we can rescale the components of $\sigma'(0)$.\footnote{Note that this does not change the parametrization of the geodesic, hence we still have $\sigma(0)=p$ and $\sigma(1)=q$.} This means we can rescale the $y_{\perp}$ coordinates arbitrarily and hence we can use this freedom to bound $S$ in $y$ coordinates by the same $\delta$ that appeared in \eqref{eq:5},
\begin{align}
  S_{y}=\left\{ (y_{\parallel},y_{\perp})\in\mathbb{R}^{N_{\mathcal{X}}}\ \Big|\ y_{\parallel}\in\mathcal{D}_{y},\ y_{\perp}^{i}\in \Big(-\frac{\delta}{2},\frac{\delta}{2}\Big)\right\}\,.\label{eq:6}
\end{align}
Furthermore, in $g(\mathcal{Z})$, we can choose the basis $\{n_{i}\}$ orthogonal with respect to the (singular) metric $\gamma$ and obtain in some neighborhood of $\mathcal{D}\cap g(\mathcal{Z})$
\begin{align}
  \gamma^{\mu\nu}(y)=\begin{pmatrix}
    \gamma^{-1}_{\mathcal{D}}(y)&&&\\
    &\gamma^{-1}_{\perp_{1}}&&\\
    &&\ddots&\\
    &&&\gamma^{-1}_{\perp_{N_{\mathcal{X}}-N_{\mathcal{D}}}}
  \end{pmatrix}^{\mu\nu}\,.\label{eq:7}
\end{align}
Note that this form of the metric together with \eqref{eq:6} implies in particular that $S$ takes the product form mentioned in \eqref{eq:prod_mfd}.
In the following, we will show that for well-trained generative networks and thin data distributions (i.e.\ for small $\delta$), $\gamma_{\perp_{i}}^{-1}\rightarrow0$. To understand the consequences for the gradient ascent update step, consider \eqref{eq:4} in $y^{\mu}$ coordinates
\begin{align}
  \gamma^{\alpha\beta}\frac{\partial f_{t}}{\partial x^{\beta}}
  &= \frac{\partial x^{\alpha}}{\partial y^{\mu}}\gamma^{\mu\nu}\frac{\partial f_{t}}{\partial y^{\nu}}\nonumber\\
  &= \frac{\partial x^{\alpha}}{\partial y_{\parallel}^{\mu}}\gamma_{\mathcal{D}}^{\mu\nu}\frac{\partial f_{t}}{\partial y_{\parallel}^{\nu}} + \frac{\partial x^{\alpha}}{\partial y_{\perp}^{i}}\gamma_{\perp_i}^{-1}\frac{\partial f_{t}}{\partial y_{\perp}^{i}}\,.
\end{align}
For $\gamma_{\perp_{i}}^{-1}\rightarrow0$ and $\frac{\partial x}{\partial y_{\perp}}$ bounded, the second term vanishes and we arrive at
\begin{align}
  \gamma^{\alpha\beta}\frac{\partial f_{t}}{\partial x^{\beta}} \rightarrow \frac{\partial x^{\alpha}}{\partial y_{\parallel}^{\mu}}\gamma_{\mathcal{D}}^{\mu\nu}\frac{\partial f_{t}}{\partial y_{\parallel}^{\nu}}\label{eq:14}
\end{align}
and hence the orthogonal directions in the update step \eqref{eq:4}, leading away from the data manifold $\mathcal{D}$, are suppressed. Therefore, \eqref{eq:4} produces counterfactuals instead of adversarial examples.

\subsection{Diffeomorphic Counterfactuals}\label{sec:flows}
In this section, we show that for well-trained normalizing flows, the orthogonal components of the inverse metric $\gamma_{\perp_i}^{-1}$ vanish for thin data manifolds, as formalized in the following theorem.
\begin{theorem}\label{th:diff_ctfctls}
    For $\epsilon\in(0,1)$ and $g$ a normalizing flow with Kullback--Leibler divergence $\KL(p,q)<\epsilon$,
    \begin{align*}
        \gamma_{\perp_i}^{-1}\rightarrow0\qquad\text{as}\qquad\delta\rightarrow0
    \end{align*}
    for all $i \in \{1,\dots,N_\mathcal{X}-N_\mathcal{D}\}$.
\end{theorem}
The main argument of the formal proof given in Appendix~\ref{app:diff_ctfctls_proof} proceeds as follows: First, we show that a small Kullback--Leibler divergence implies that most of the induced probability mass lies in the support of the data distribution,
\begin{align}
  \int_{S_{x}}q_{x}(x)\dd{x}>1-\epsilon\,.\label{eq:17}
\end{align}
Next, we write $q_{x}$ as the pull-back of the latent distribution $q_{z}$ under the flow $g$ using the familiar change-of-variables formula for normalizing flows. In the $y^\mu$ coordinates introduced above, the resulting integral then factorizes according to the block-diagonal structure \eqref{eq:7} of the metric with integration domain $[-\delta/2,\delta/2]$ for the $y_\perp^i$ directions. As $\delta\rightarrow0$, the bound \eqref{eq:17} can only remain satisfied if the associated metric component $\gamma_{\perp_i}$ diverges, implying that $\gamma_{\perp_i}^{-1}\rightarrow0$.

Following the steps at the end of Section~\ref{sec:math-setup}, we see that this necessarily implies that the gradient ascent update \eqref{eq:1} stays on the data manifold, since $\frac{\partial x}{\partial y_{\perp}}$ is constant (and therefore bounded) as $\delta\rightarrow0$.

\subsection{Approximate Diffeomorphic Counterfactuals}
\label{sec:gener-gener-netw}

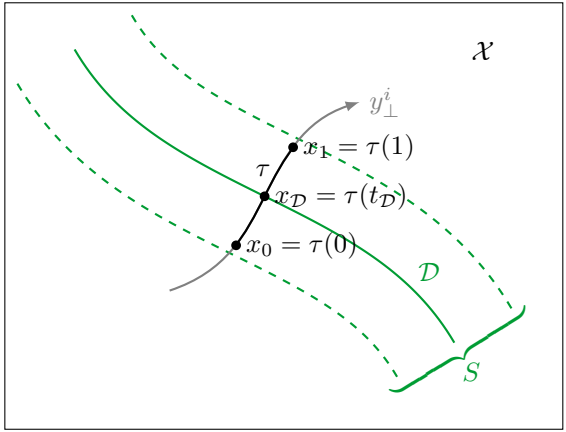
\begin{figure}[t]
  \centering
  \begin{tikzpicture}[scale=2.5,show background rectangle, line width=0.8pt]
    \draw[postaction={decoration={contour lineto, contour distance=-25},draw=MyGreen, dashed, decorate},
          postaction={decoration={contour lineto, contour distance=25},draw=MyGreen, dashed, decorate},
          domain=2.1416:4.1416,samples=100,MyGreen] plot (\x, {0.5*tan(-\x r)});
    \node at (4.23,-0.98) [MyGreen,anchor=base] {$S$};
    \node[MyGreen,rotate=-58,scale=1] at (4.18,-0.82) {$\left.\vphantom{\raisebox{3em}{x}}\right\}$};
    \node[black] at (4.2916,0.77) {$\mathcal{X}$};
    \node[MyGreen,anchor=west] at (3.9,-0.4) {$\mathcal{D}$};
    \begin{scope}
      \begin{scope}[shift={(3.1416,0)}]
        \draw[->,>=latex,gray] (-0.5,-0.5) .. controls (-0.2,-0.4) and (-0.1,-0.2) ..
        (0,0) .. controls (0.1,0.2) and (0.2,0.4) .. (0.5,0.5) node[anchor=west]{$y_{\perp}^{i}$};
        \draw[fill] (0,0) circle (0.02) node[anchor=west]{$x_{\mathcal{D}}=\tau(t_{\mathcal{D}})$};
        \draw[fill] (0.15,0.26) circle (0.02) node[anchor=west]{$x_{1}=\tau(1)$};
        \draw[fill] (-0.15,-0.26) circle (0.02) node[anchor=west]{$x_{0}=\tau(0)$};
      \end{scope}
      \begin{scope}[shift={(3.1416,0)}]
        \clip (-0.2,0.26) -- (0.2,0.26) -- (0.2,-0.26) -- (-0.2,-0.26);
        \draw[black] (-0.5,-0.5) .. controls (-0.2,-0.4) and (-0.1,-0.2) ..
        (0,0) node[anchor=east,yshift=10,xshift=6] {$\tau$} .. controls (0.1,0.2) and (0.2,0.4) .. (0.5,0.5);
      \end{scope}
    \end{scope}
  \end{tikzpicture}
  \caption{Construction of the curve $\tau$ used in Section~\ref{sec:gener-gener-netw}.}
  \label{fig:tau}
\end{figure}

We now present a theorem similar to Theorem~\ref{th:diff_ctfctls} for the case of approximately diffeomorphic counterfactuals, i.e.\ for AEs and GANs, showing that these models can also be used to construct counterfactuals. This will however necessitate stronger assumptions since the generative model is in this case not bijective. In particular, we will assume that the generative model captures all of the data, i.e.\ that $\mathcal{D}\subset g(\mathcal{Z})$, implying that in $y$ coordinates, although $\gamma$ is singular for $N_{\mathcal{Z}}<N_{\mathcal{X}}$, the component $\gamma_{\mathcal{D}}$ is non-singular. Therefore, we split the $y_{\perp,i}$ directions into $N_{\mathcal{X}}-N_{\mathcal{Z}}$ singular directions and $N_{\mathcal{Z}}-N_{\mathcal{D}}$ non-singular directions. Since the inverse metric vanishes by definition in the singular directions, the theorem focuses on the non-singular directions and can then be stated as follows,

\begin{theorem}\label{th:approx_diff_ctfctls}
    If $g:\mathcal{Z}\rightarrow\mathcal{X}$ is a generative model with $\mathcal{D}\subset g(\mathcal{Z})$ and image $g(\mathcal{Z})$ which extends in any non-singular orthogonal direction $y_{\perp}^{i}$ outside of $\mathcal{D}$,
    \begin{align*}
        \gamma_{\perp_{i}}^{-1}\rightarrow 0
    \end{align*}
    for $\delta\rightarrow0$ for all non-singular orthogonal directions $y_{\perp}^{i}$.
\end{theorem}
The proof can be found in Appendix~\ref{app:proof_approx_diff_ctfctls_proof} and proceeds as follows: First, we construct a curve $\tau:[0,1]\rightarrow\mathcal{Z}$ which cuts through $S$ along the $y^i_\perp$-coordinate line and lies completely in $g(\mathcal{Z})$, as illustrated in Figure~\ref{fig:tau}. Then, the length $\mathcal{L}(\tau)$ of this curve (with respect to $\gamma$) computed in $y^\mu$-coordinates is, for small $\delta$, approximately given by 
\begin{align}
    \mathcal{L}(\tau)\approx\sqrt{\gamma_{\perp_{i}}(x_{\mathcal{D}})}\,(x_{1,\perp}{}^{i}-x_{0,\perp}{}^{i})\,.
\end{align}
Bounding the difference by $\delta$ and using that $\mathcal{L}(\tau)$ is constant, yields the desired result. As in the case of Theorem~\ref{th:diff_ctfctls} above, this implies again that the gradient ascent update \eqref{eq:1} does not leave the data manifold as shown in \eqref{eq:14}.

\section{Experiments}\label{sec:experiments}
Equipped with our theoretical results, we are now ready to  present our experimental findings. 

We start by illustrating diffeomorphic explanations using a toy example in three-dimensional space. This allows us to directly visualize the data manifold and the trajectories of gradient ascent in $\mathcal{X}$ and $\mathcal{Z}$.

We then apply our diffeomorphic counterfactual method, using normalizing flows, to four different image data sets. We use MNIST, CelebA and CheXpert for classification tasks and the Mall data set for a regression task. We evaluate the results qualitatively and quantitatively. 
Furthermore we discuss approximate diffeomorphic counterfactuals, using VAEs and GANs, which allow us to consider high resolution data.

For all experiments, we use the same setup:
We require a pretrained generator $g$ and a pretrained classifier $f$. We start with a data point $x$ from the test set that is predicted by the classifier $f$ as belonging to the source class. We define target class $t$ and target confidence $\Lambda$. To produce an adversarial example, we then update the original data point following the gradient in $\mathcal{X}$, $\frac{\partial f_t(x)}{\partial x}$, until we reach the desired target confidence. To produce a counterfactual we first project the original data point into the latent space of the generative model $g$ by applying the inverse generative model $g^{-1}(x)=z$, or an appropriate approximation (for GANs). We then update the original latent representation $z$ following the gradient in $\mathcal{Z}$, $\frac{\partial (f_t\circ g) (z)}{\partial z}$, until we reach the desired target confidence.

For more details on model configuration, training and hyperparameters we refer to Appendix~\ref{app:experiments}.

\subsection{Diffeomorphic Counterfactuals}\label{sec:experiments_diffeo}

\subsubsection{Toy example}\label{sec:toy_example}
\begin{figure}[t]
\centering
\includegraphics[width=1\linewidth]{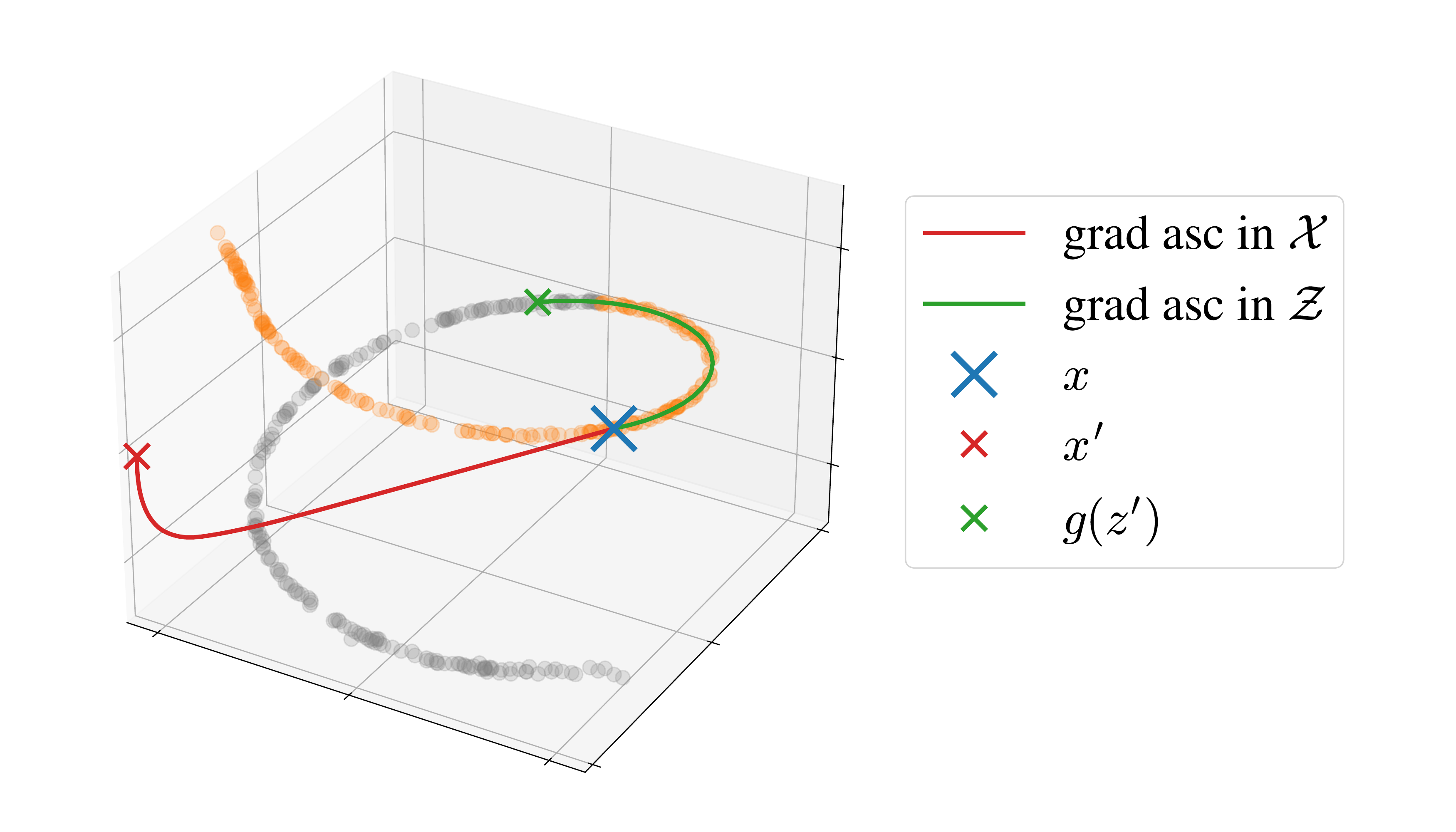}
 \caption{Gradient ascent in $\mathcal{X}$ leads to points that lie significantly off-manifold while gradient ascent in $\mathcal{Z}$ moves along the data manifold. The ground truth for different classes is depicted in orange (source class) and gray (target class).}
\label{fig:toy_adv_attack}
\end{figure}
We consider data uniformly distributed on a one-dimensional manifold, a helix, that is embedded in three-dimensional space and train a simple normalizing flow that approximates the data distribution.
As illustrated in Figure~\ref{fig:toy_adv_attack}, we divide the data into two classes corresponding to the upper and the lower half of the helix and train a classifier.

We then generate counterfactuals by the gradient ascent optimization in input space $\mathcal{X}$ and in the latent space of the flow $\mathcal{Z}$, i.e. by using \eqref{eq:2} and \eqref{eq:1} respectively.

Starting from the original data point $x$, we observe that gradient ascent in $\mathcal{X}$ leads to points that lie significantly off data manifold $S$. In contrast to that, the updates of gradient ascent in the latent space $\mathcal{Z}$ follow a trajectory along the data manifold resulting in counterfactuals with the desired target classification which lie on the data manifold. We illustrate this in Figure~\ref{fig:toy_adv_attack}.

As we have an analytic description of the data manifold, we can reliably calculate the distances to the data manifold for all points found via gradient ascent in $\mathcal{X}$ or $\mathcal{Z}$.
We compare 1000 successful optimizations (all optimizations reach the desired target confidence) in the input space $\mathcal{X}$ and latent space $\mathcal{Z}$. The median value for the distances to the data manifold when performing gradient ascent in $\mathcal{X}$ or in $\mathcal{Z}$ are $2.34$ and $0.01$ respectively (see also Figure~\ref{fig:toy_quantitative} in the appendix). This intuitively and clearly illustrates the benefit of performing gradient ascent in the latent space $\mathcal{Z}$.

\subsubsection{Tangent space of data manifold}
\begin{figure}[t]
\begin{center}
\centerline{\includegraphics[height=0.4\columnwidth]{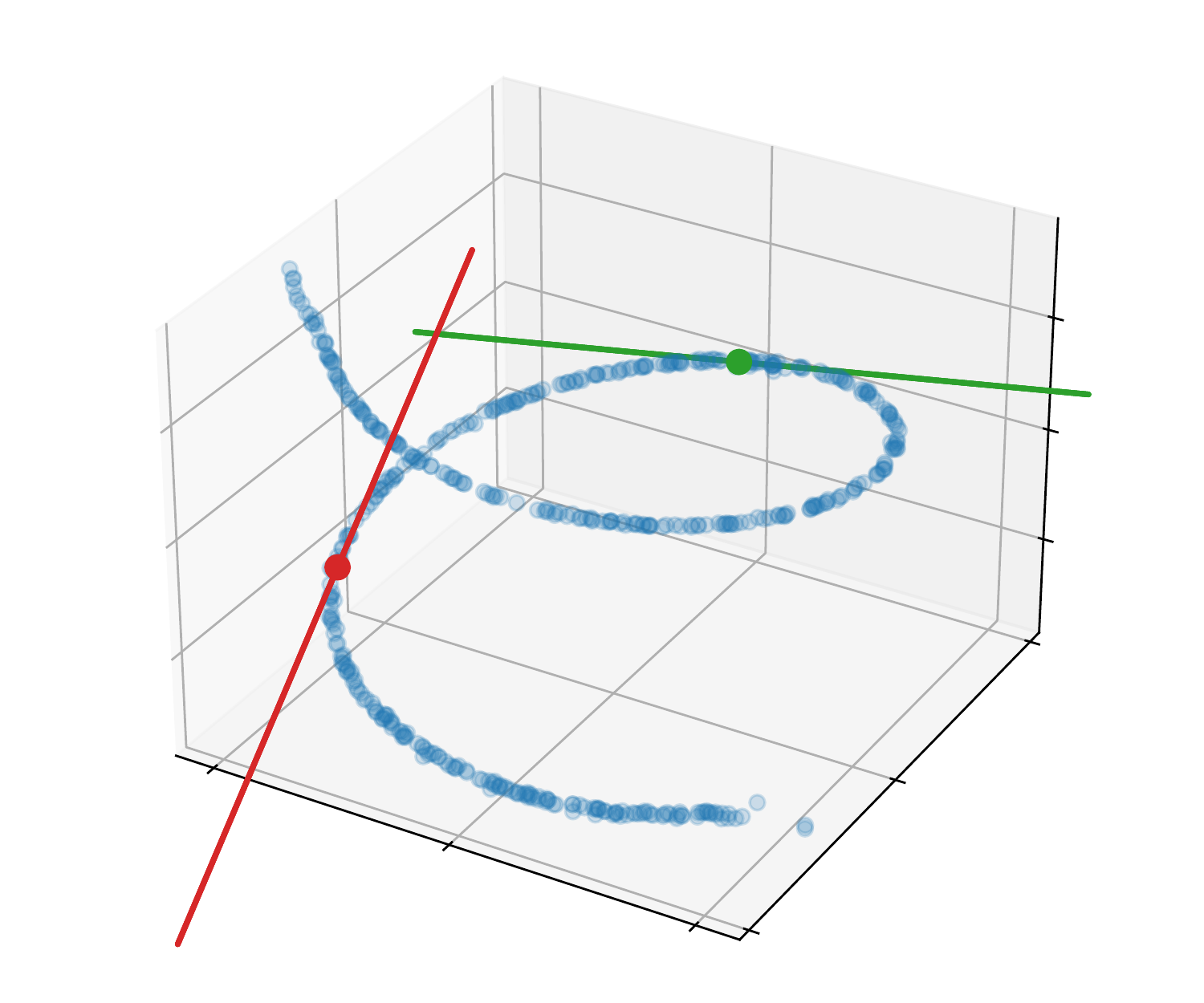}\hspace{-0.3cm}
\includegraphics[height=0.35\columnwidth]{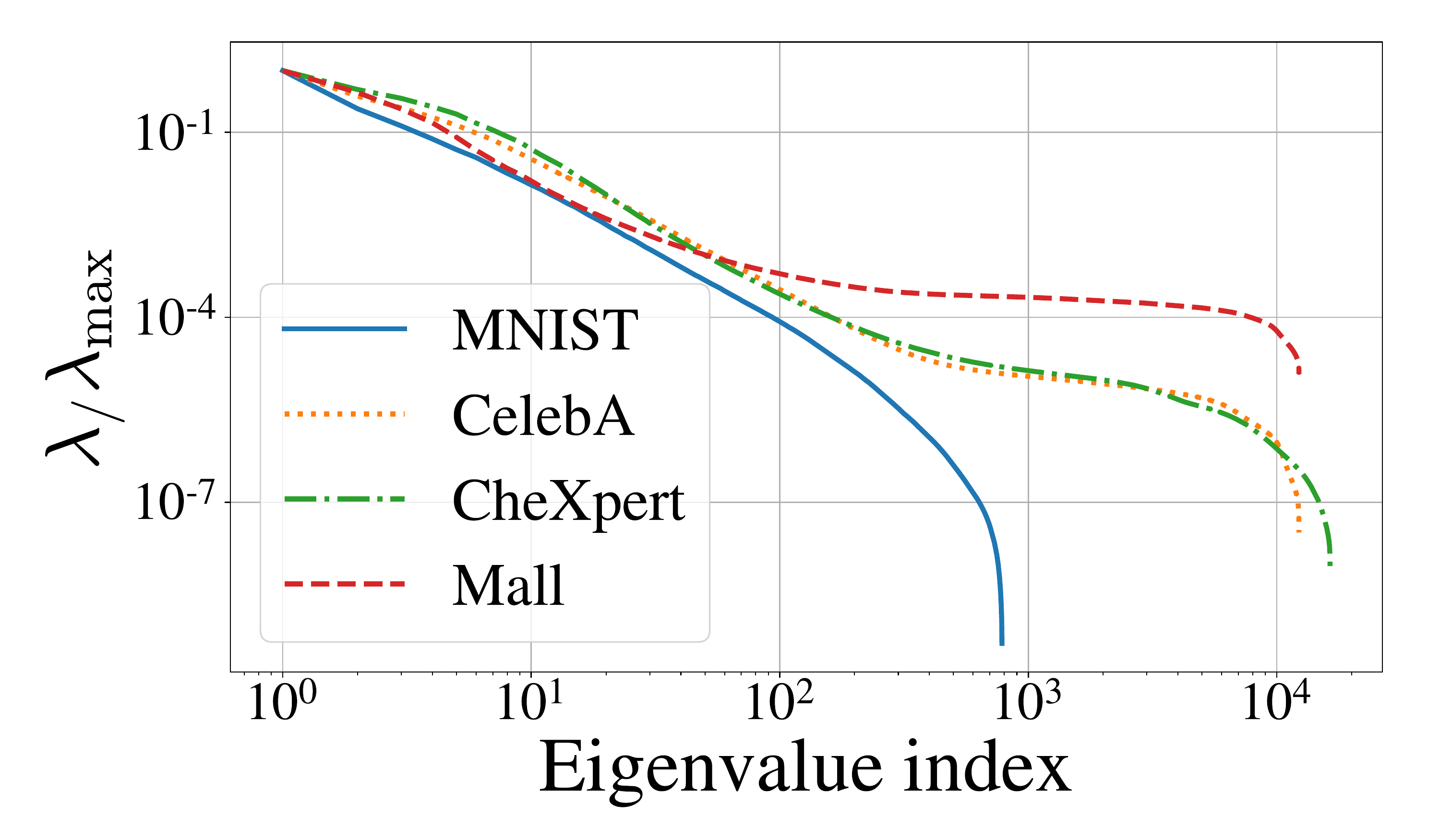}}
\caption{\textbf{Left:} As expected from the theoretical analysis, the parallelepiped spanned by
all three eigenvectors of the inverse induced metric scaled by the
corresponding eigenvalues is to good approximation one-dimensional, i.e. of the same
dimension as the data manifold, and tangential to it. \textbf{Right:} The Jacobians of the trained flows have a low number of large and a large number of small eigenvalues, suggesting that the images lie approximately on a low-dimensional manifold. Both axes are scaled logarithmically.}
\label{fig:eigen_values}
\end{center}
\vskip -0.2in
\end{figure}

A non-trivial consequence of our theoretical insights is that we can infer the tangent space of each point on the data manifold from our flow $g$. Specifically, we perform  a singular value decomposition of the Jacobian $\frac{\partial g}{\partial z} = U \, \Sigma \, V$ and rewrite the inverse induced metric as
\begin{align}
    \gamma^{-1} = \frac{\partial g}{\partial z} \frac{\partial g}{\partial z}^T = U \, \Sigma^2 \, U^T \,.
\end{align}
As we saw in Section~\ref{sec:theory}, for data concentrated on an $N_{\mathcal{D}}$-dimensional data manifold $\mathcal{D}$ in an $N_{\mathcal{X}}$-dimensional embedding space $\mathcal{X}$, the inverse induced metric $\gamma^{-1}$ has $N_{\mathcal{X}}-N_{\mathcal{D}}$ small eigenvalues. Furthermore, the eigenvectors corresponding to the large eigenvalues will approximately span the tangent space of the data manifold. 
For our toy example from Section~\ref{sec:toy_example}, we can directly show the parallelepiped spanned by the three eigenvectors in three-dimensional space. Figure~\ref{fig:eigen_values} (left) indeed shows that the parallelepipeds are significantly contracted in two of the three dimensions making them appear as one dimensional lines. For the high dimensional image data sets, which are discussed in Section~\ref{sec:experiments_images}, we show the sorted eigenvalues, averaged over 100 random data points per data set, cf.\  Figure~\ref{fig:eigen_values} (right). Our experiments confirm the theoretical expectation that the large eigenvectors indeed span the tangent space of the manifold.

\subsubsection{Image classification and regression}\label{sec:experiments_images}
\begin{figure*}[!t]
\begin{center}
\centerline{\includegraphics[width=1.\linewidth]{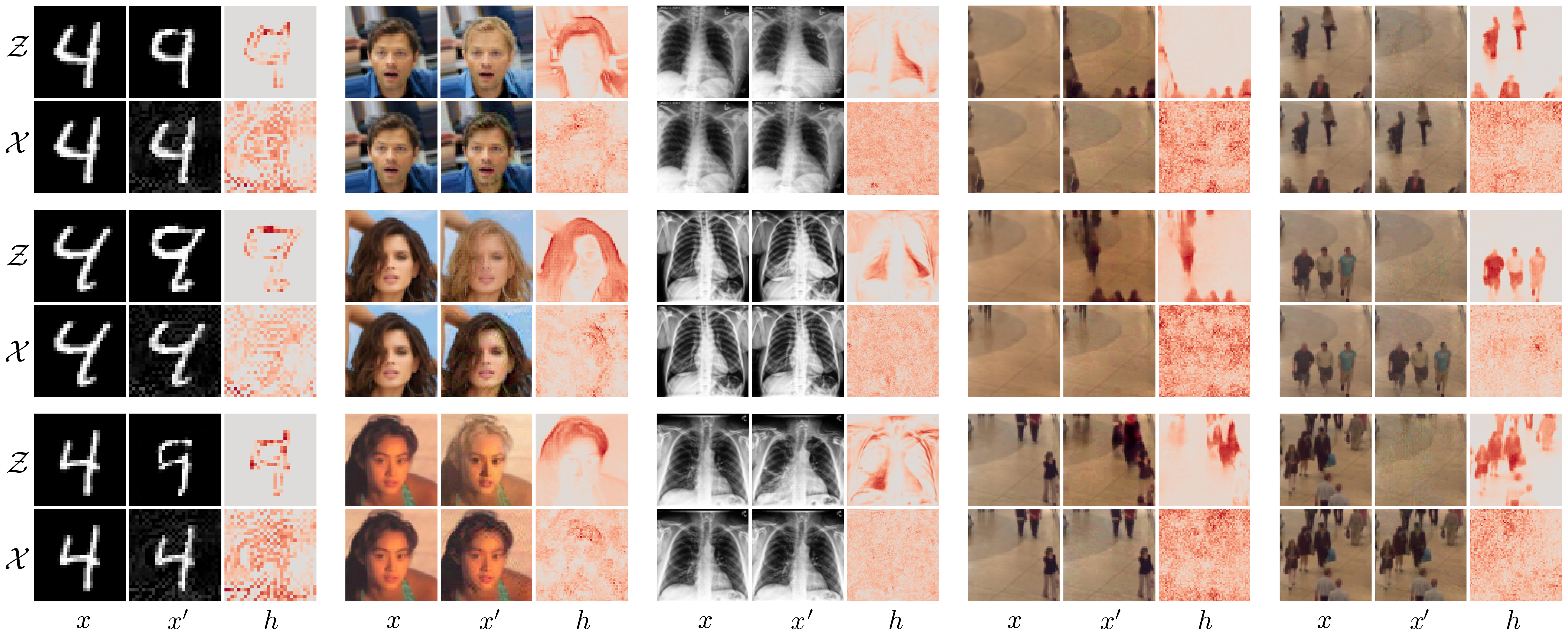}}
\caption{Counterfactuals for MNIST (`four' to `nine'), CelebA (`not-blond' to `blond'), CheXpert (`healthy' to `cardiomegaly'), Mall (`few' to `many') and Mall (`many' to `few'). Columns of each block show original image $x$, counterfactual $x'$, and difference $h$ for three selected datapoints. First row of each block is our diffeomorphic counterfactuals, i.e. obtained by gradient ascent in $\mathcal{Z}$ space. Second row of each block is standard gradient ascent in $\mathcal{X}$ space. Heatmaps $h$ show the difference $|x - x'|$ summed over color channels.}
\label{fig:adv_attacks}
\end{center} 
\vskip -0.2in
\end{figure*}

We now demonstrate applications of diffeomorphic counterfactuals to image classification in several domains.
\\
\\
\textbf{\underline{Datasets:}} We use MNIST~ \cite{deng2012mnist}, CelebA~\cite{liu2015faceattributes}, CheXpert~\cite{irvin2019chexpert} (a data set of labeled chest X-rays) and a the Mall data set~\cite{chen2012feature, chen2013cumulative, change2013semi, loy2013crowd} (a crowd-counting data set with video frames from a shopping mall with head annotations of pedestrians).
\\
\\
\textbf{\underline{Classifiers/Regressors:}} We train a ten-class CNN on MNIST (test accuracy of 99\%). For CelebA, we train a binary CNN on the blond attribute (test accuracy of 94\%). For CheXpert, we train a binary CNN on the cardiomegaly attribute (test accuracy of 86\%). 

For the Mall data set, we train a U-Net~\cite{ronneberger2015unet} that outputs a probability map of the size of the image and a scalar regression value, which corresponds to the approximated number of pedestrians in the picture. Following the definitions by Ribera et al.~\cite{ribera2019locating}, our trained U-Net reaches a RMSE for the head count of 0.63. When we run our gradient ascent algorithm, we aim to maximize/minimize merely the scalar regression value, i.e. the number of pedestrians.  
\\
\\
\textbf{\underline{Flows:}} We choose a flow with RealNVP-type couplings \cite{dinh2016density} for MNIST and the Glow architecture \cite{glow} for CelebA, CheXpert and the Mall data set. 
\\
\\
\textbf{\underline{Generation of counterfactuals:}} We start from original data points $x$ of the classes `four' for MNIST, `not blond' for CelebA and `healthy' for CheXpert. We select the classes `nine', `blond', and `cardiomegaly' as targets $t$ for MNIST, CelebA, and CheXpert, respectively, and take the confidence threshold to be $\Lambda=0.99$. For the Mall data set, we maximize the regression value $r$ (threshold at $r=10$) if few pedestrians were detected in the original image $x$ and minimize the regression value (threshold at $r=0.01$) if many pedestrians were detected in the original image $x$. We use Adam to optimize in $\mathcal{X}$ or $\mathcal{Z}$ until the confidence threshold $\Lambda$ for the target class $t$ is reached.
\\
\\
\textbf{\underline{Qualitative analysis:}} Our diffeomorphic counterfactuals produced by the normalizing flows indeed show semantically meaningful deformations in particular when compared to adversarial examples produced by gradient ascent in the data space $\mathcal{X}$. 

We show examples in Figure~\ref{fig:adv_attacks}. The counterfactuals resemble  images from the data set that have the target class as the ground truth label. At the same time the counterfactuals are similar to their respective source images with respect to features that are irrelevant for the differentiation between source and target class. 

For MNIST, the stroke width and the writing angle remain unchanged in the counterfactuals while the gap in the upper part of the 'four' changes to the characteristic upper loop of the 'nine'. 

For CelebA, the changes in the counterfactuals are focused on the hair area as evident from the heatmaps. Facial features and background stay (approximately) constant. 

The counterfactuals for the CheXpert data set mostly brighten the pixels in the central region of the picture leading to the appearance of an enlarged heart. The other structures in the image remain mostly constant.

Also for pictures taken from the Mall data set, we observe that the counterfactuals remain close to the original images. When maximizing the regression value, pedestrians are generated at the picture's edge or appear around darker areas in the original image. When minimizing pedestrians, we observe that the counterfactuals reproduce the darker parts of the floor and lines between the tiles.
\\
\\
\textbf{\underline{Quantitative analysis:}} To quantitatively assess the quality of our counterfactuals, we use several measures, as detailed in the following.
\\
\\
\underline{Oracle:} We train a 10-class SVM on MNIST (test accuracy of 92\%) and binary SVMs on CelebA (test accuracy of 85\%) and CheXpert (test accuracy of 70\%). The counterfactuals found by performing gradient ascent in the base space of the flow generalize significantly better to these simple models suggesting that they indeed use semantically more relevant deformations than conventional adversarial examples produced by gradient ascent in $X$ space. 
 
\begin{figure}[t]
\begin{center}
\centerline{\includegraphics[height=0.4\columnwidth]{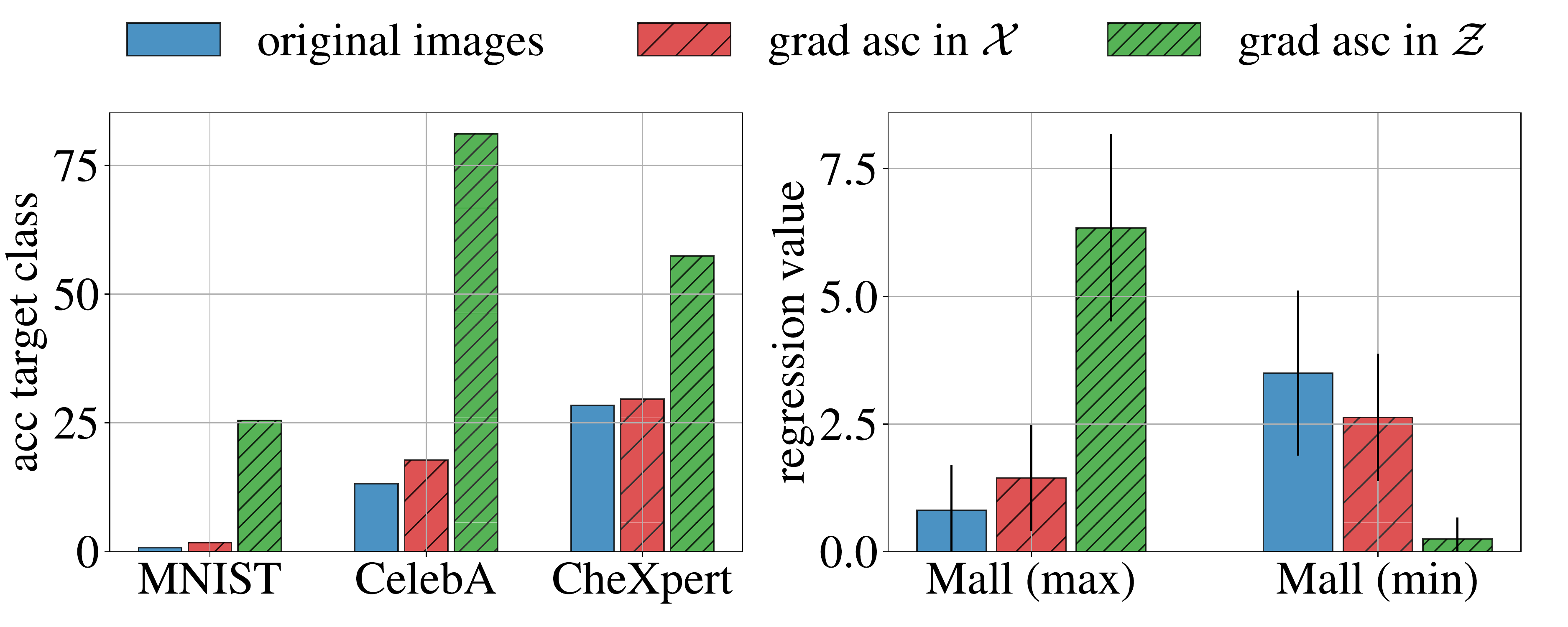}}
\caption{\textbf{Left:} accuracy with respect to the target class $k$ generalizes better to SVM for diffeomorphic counterfactuals. \textbf{Right:} regression values for oracle are closer to target values for $\mathcal{Z}$-based counterfactuals (bars show means and errors denote one standard deviation).}
\label{fig:quantitative_oracle}
\end{center}
\vskip -0.2in
\end{figure}

For the Mall data set, we train a slightly larger U-Net (RMSE for head count 0.72) and calculate regression values for the original images, the images modified with gradient ascent in $\mathcal{X}$-space and the images modified with gradient ascent in $\mathcal{Z}$-space. As expected, the regression values for the counterfactuals are significantly closer to the target values (0.01 for minimizing pedestrians and 10 for maximizing pedestrians) than those of original images and adversarial examples. 
Figure~\ref{fig:quantitative_oracle} summarizes these findings.

\begin{figure}[t]
\begin{center}
\centerline{\includegraphics[width=.8\columnwidth]{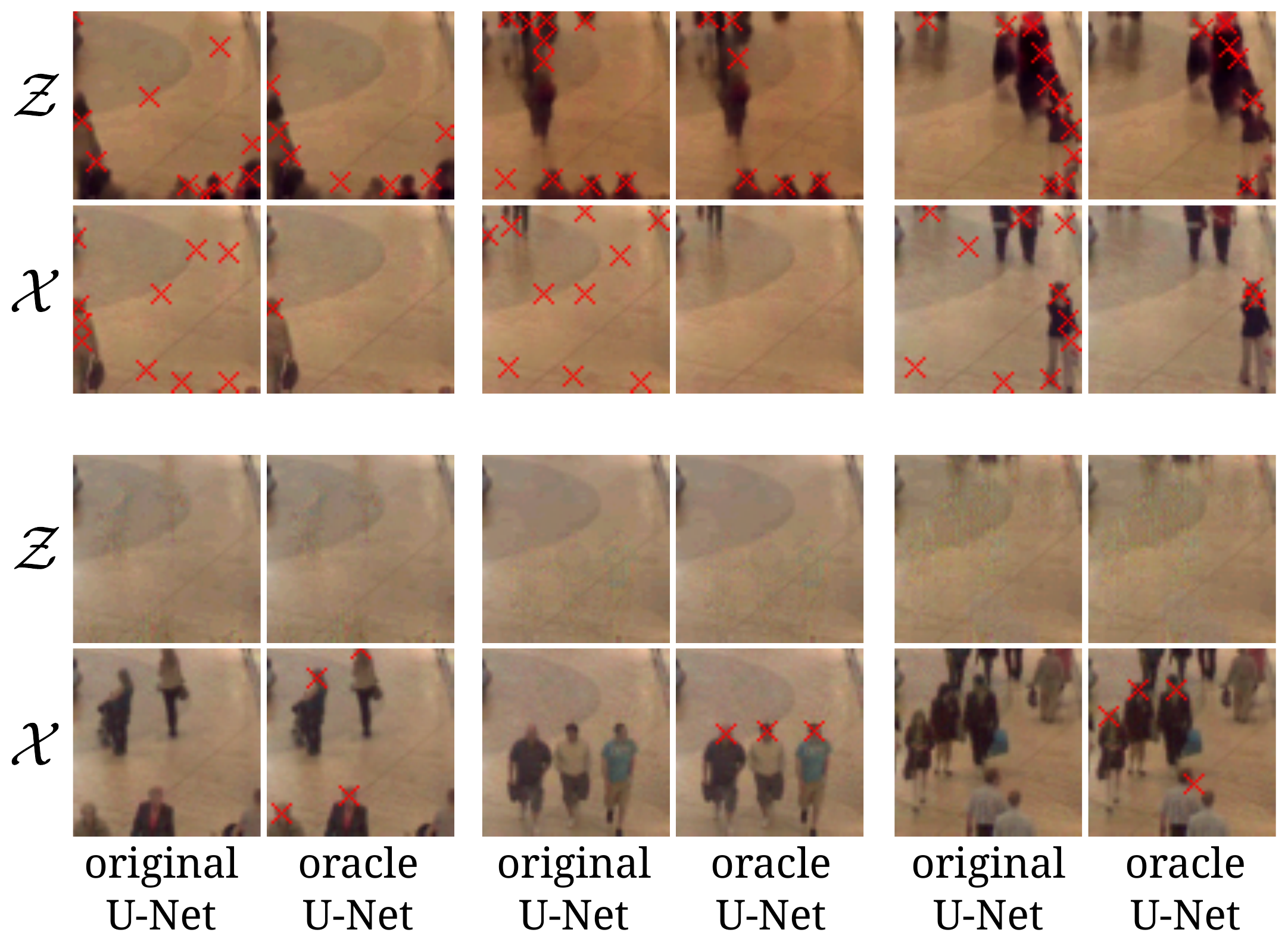}}
\caption{Head locations for pedestrians in counterfactuals and adversarial examples when maximizing pedestrians (upper two rows) and minimizing pedestrians (lower two rows). The original U-Net is fooled by the adversarial examples, leading to false positives (second row) and false negatives (forth row) when detecting pedestrians. The oracle U-Net generalizes to the diffeomorphic counterfactuals found by gradient ascent in $\mathcal{Z}$ (odd rows) but not to the adversarial examples found by gradient ascent in $\mathcal{X}$ (even rows).}
\label{fig:head_locations}
\end{center}
\vskip -0.2in
\end{figure}

\begin{figure}[t]
\begin{center}
\centerline{\includegraphics[width=1.0\columnwidth]{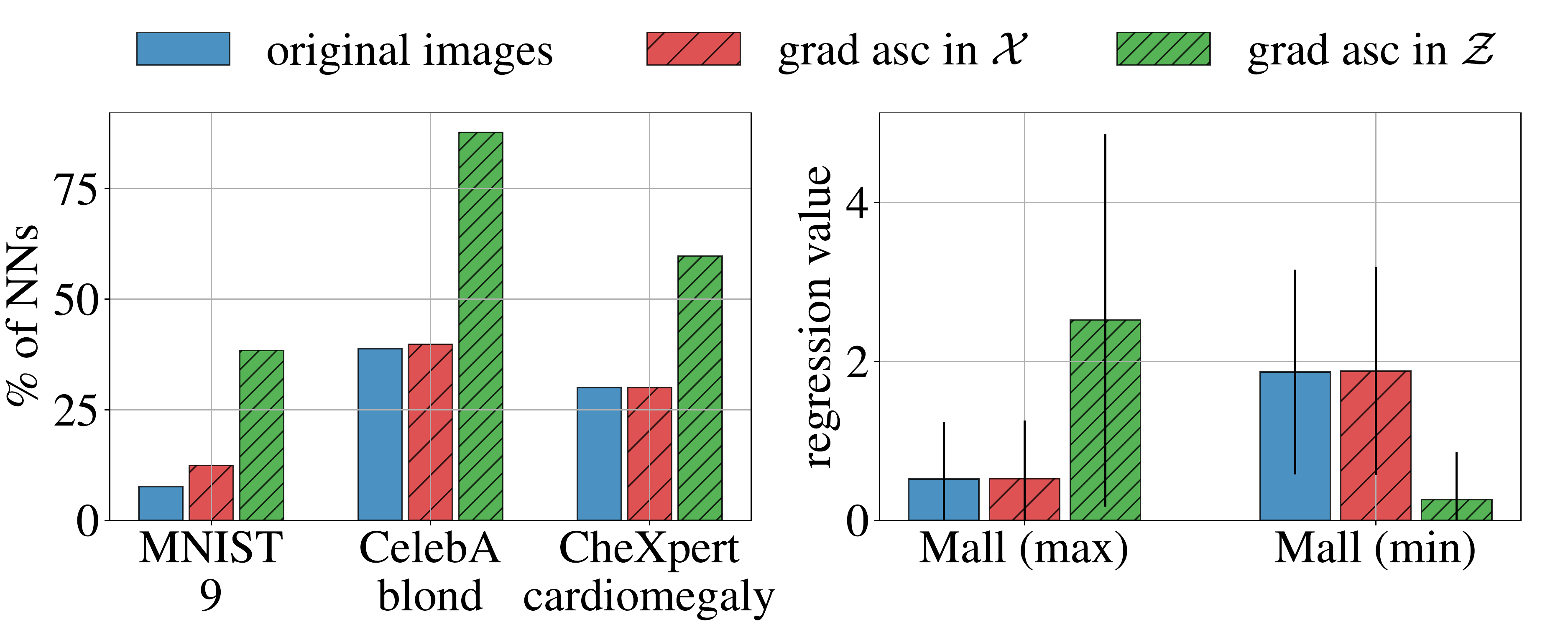}}
\caption{\textbf{Left:} ground truth class for the ten nearest neighbours (NNs) matches the target value (`9', `blond' and `cardiomegaly') more often for the counterfactuals found in $\mathcal{Z}$. \textbf{Right:} ground truth pedestrian counts averaged over the three nearest neighbours are closer to target values for diffeomorphic counterfactuals. Bars show means and errors denote one standard deviation.}
\label{fig:quantitative_prototypes}
\end{center}
\vskip -0.2in
\end{figure}

In Figure~\ref{fig:head_locations}, we show the localization of heads for the counterfactuals and the adversarial examples for the Mall data set from Figure~\ref{fig:adv_attacks} using the original and the oracle U-Net.
In order to find the head locations, the regression value is rounded to the closest integer representing the number of pedestrians in the image. A Gaussian mixture model with the number of pedestrians as components is then fitted to the probability map. Finally the head positions are defined as the means of the fitted Gaussians. 
The original U-Net is deceived by the adversarial examples: When maximizing pedestrians (second row) the original U-Net produces false positives, leading to markers at head locations where there are no pedestrians. When minimizing pedestrians, the adversarial examples (forth row) fool the original U-Net into making false negative errors, that is failing to detect pedestrians, although they are clearly present. The oracle U-Net on the other hand produces regression values and probability maps that enable correct identification of pedestrian's head positions (or lack thereof) for the adversarial examples when maximizing (second row) and minimizing (forth row) pedestrians. For the diffeomorphic counterfactuals (first and third row in Figure~\ref{fig:head_locations}), the predictions of the two U-Nets are similar, showing that these counterfactuals generalize to the independently trained oracle U-Net.
\\
\\
\underline{Nearest neighbours:} We compare the original images and the images modified in $\mathcal{X}$ and $\mathcal{Z}$ with data from the data set. We find the $k$-nearest neighbours (with respect to the Euclidean norm) and their respective ground truth classification/regression value. For MNIST, CelebA and CheXpert, we then check what percentage of the nearest neighbours was classified as the target class. For Mall, we check the average number of pedestrians present. Figure~\ref{fig:quantitative_prototypes} shows that the ten nearest neighbours of the diffeomorphic counterfactuals for MNIST, CelebA and CheXpert share the target classification more often than original images or adversarial examples. For the Mall data set the three nearest neighbours of each counterfactual on average have regression values that more closely match the target regression value ($r=10$ when maximizing and $r=0$ when  minimizing).
\\
\\
\underline{IM1 and IM2:} Van Looveren and Klaise~\cite{looveren2021interpretable} propose two metrics to test interpretability: IM1 is defined by
\begin{equation}
    \text{IM1} = \frac{||x'-\text{AE}_t(x')||}{||x'-\text{AE}_{c_0}(x')||+\epsilon}\,,\label{eq:12}
\end{equation}
where $\text{AE}_{c_0 }$ and $\text{AE}_t$ are two autoencoders which were each trained on data from only one class (original class $c_0$ and target class $t$, respectively) and $\epsilon$ is a small positive value. The second metric IM2 is defined by
\begin{equation}
    \text{IM2}= \frac{||\text{AE}_t(x')-\text{AE}(x')||}{||x'||_1 + \epsilon}\,,\label{eq:13}
\end{equation}
where $\text{AE}$ is an autoencoder trained on all classes.

\begin{figure}[t]
\begin{center}
\centerline{\includegraphics[width=1.0\columnwidth]{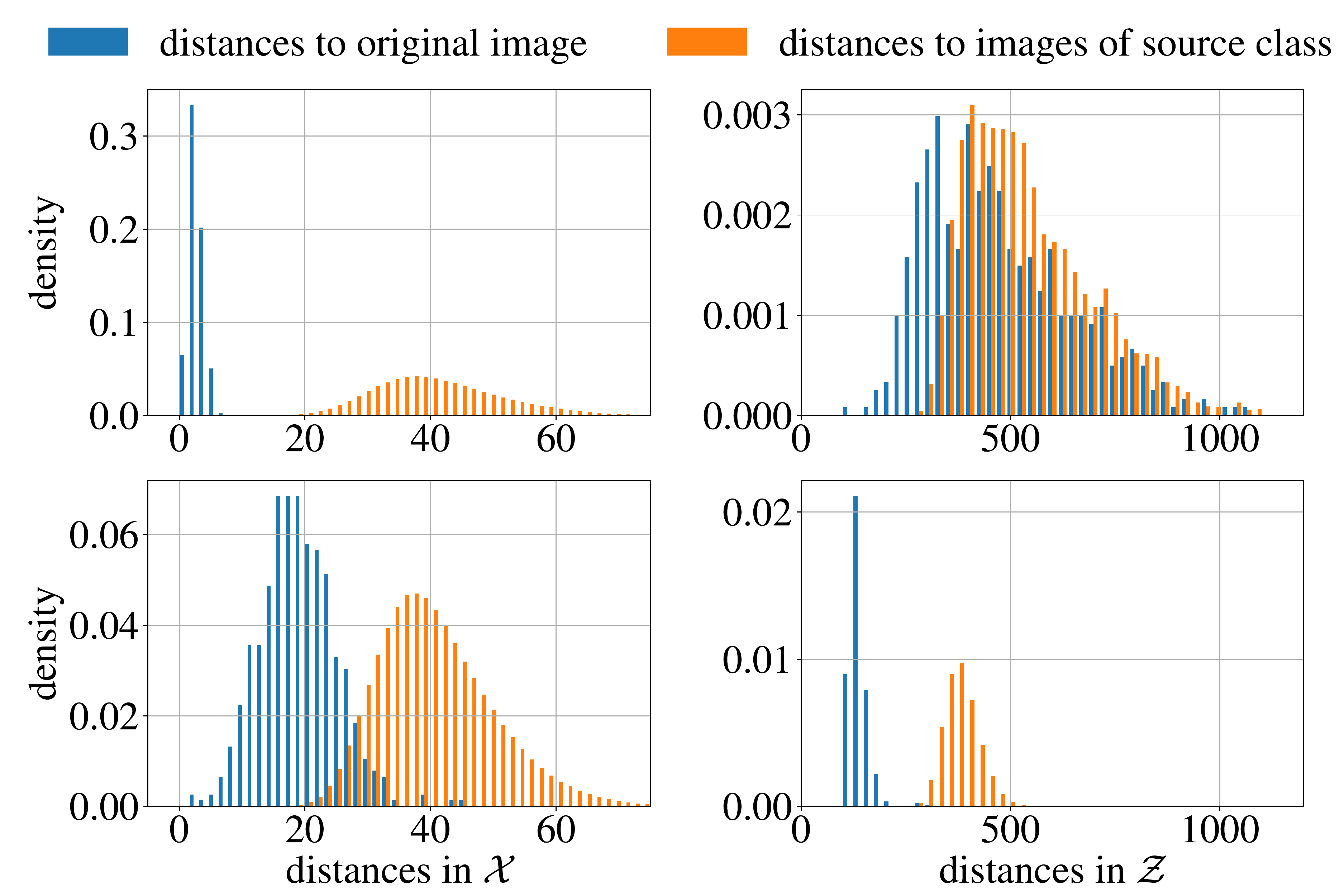}}
\caption{Euclidean distances in $\mathcal{X}$ and $\mathcal{Z}$ for adversarial examples (first row) and counterfactuals (second row) for the CelebA dataset.
Counterfactuals lie closer to their respective source image than adversarial examples when measured in $\mathcal{Z}$, i.e. along the data manifold.}
\label{fig:L2_source_CelebA}
\end{center}
\vskip -0.2in
\end{figure}

IM1 and especially IM2 have been repeatedly criticized~\cite{hvilshoj2021quantitative, mahajan2019preserving, schut2021generating}. For IM2, we devide by the one-norm $||x'||_1$ of the modified image. This value is large if the image has more bright pixels. Consequently, images with brighter pixels will tend to have a smaller IM2, even though they might not be more interpretable. We therefore limit our evaluation to IM1.
\begin{table}[htbp]
\centering
\footnotesize
\begin{tabular}{|c|c|c|}
\hline
data set & images & IM1 \\ \hhline{|=|=|=|}
\multirow{3}{*}{MNIST} & original & 2.250 $\pm$ 0.711 \\
& gradient ascent in $\mathcal{X}$ & 1.603 $\pm$ 0.317 \\
& gradient ascent in $\mathcal{Z}$ & 1.056 $\pm$ 0.233 \\ \hline
\multirow{3}{*}{CelebA} & original &  1.160 $\pm$ 0.303 \\
& gradient ascent in $\mathcal{X}$ &  1.144 $\pm$ 0.287 \\
& gradient ascent in $\mathcal{Z}$ & 0.807 $\pm$ 0.222 \\ \hline
\end{tabular}
\caption{Interpretability metric IM1 values for MNIST and CelebA calculated for original images, adversarial examples and counterfactuals. Low values mean better interpretability. We show mean and standard deviation.}
\label{tab:IM1}
\end{table}
In Table~\ref{tab:IM1}, we show mean and standard deviation for the interpretability metric IM1 for two data sets; MNIST and CelebA. We calculate the values for the original images, the adversarial examples, produced by gradient ascent in $\mathcal{X}$ space, and the diffeomorphic counterfactuals, produced by gradient ascent in $\mathcal{Z}$ space. A low value for IM1 means the image is better represented by an autoencoder trained on only the target class. Diffeomorphic counterfactuals achieve a lower IM1 value than the adversarial examples, suggesting they are more interpretable. 
\\
\\
\noindent\underline{Similarity to original images:}
Counterfactuals are usually required to be minimal, that is they should be the closest point to the original data point, that lies on the data manifold and reaches the desired confidence $\Lambda$ with respect to the target class $t$. Unlike other approaches, we do not encourage similarity by explicitly minimizing the Euclidean norm between counterfactual and original image in $\mathcal{X}$ space since the relevant distance is to be computed by the induced metric on the data manifold $S$ or, equivalently, the flat metric in the latent space $\mathcal{Z}$. However, our counterfactuals still preserve high similarity to the respective source image. We confirm this by calculating the Euclidean distances
in $\mathcal{X}$ and $\mathcal{Z}$ between counterfactuals and all images of the source class (this effect is illustrated for the CelebA dataset in Figure~\ref{fig:L2_source_CelebA}). 

The average Euclidean norm between counterfactuals and the respective source images is significantly lower than the average Euclidean norm between counterfactuals and all images of the source class. For adversarial examples, we expect the Euclidean distances in $\mathcal{X}$ to the respective source image to be very small while the Euclidean distances in $\mathcal{Z}$ should be larger. 
Figure~\ref{fig:L2_source_CelebA} shows the distribution of distances in $\mathcal{X}$ and $\mathcal{Z}$ between counterfactuals/adversarials and their respective source images as well as distances between counterfactuals/adversarials and all images of the source class for the CelebA data set.

We refer to the Appendix~\ref{app:experiments_L2_source_images} for graphs for the other data sets.

In Table~\ref{tab:L2_source_images_X}, and Table~\ref{tab:L2_source_images_Z} we show the averaged Euclidean norms of the distances in $\mathcal{X}$ and $\mathcal{Z}$ for counterfactuals and adversarials respectively, confirming our expectiations.

\begin{table}[htbp]
\centering
\footnotesize
\begin{tabular}{|c|c|d{3.2}p{0.1cm}d{3.2}|d{3.2}p{0.1cm}d{3.2}|}
\hline
data set & img & \multicolumn{3}{c|}{$L^2$ source image}  & \multicolumn{3}{c|}{$L^2$ source class}  \\ \hhline{|=|=|===|===|}
\multirow{2}{*}{MNIST} 
& in $\mathcal{X}$ & 2.54  &$\pm$&  0.61 & 8.77 &$\pm$&  1.25 \\
& in $\mathcal{Z}$ & 4.82 &$\pm$&  1.20 & 9.27 &$\pm$&  1.32 \\ \hline
\multirow{2}{*}{CelebA} 
& in $\mathcal{X}$ & 2.84 &$\pm$&  1.15 & 41.13 &$\pm$&  10.03 \\
& in $\mathcal{Z}$ & 19.10 &$\pm$&  6.08 & 40.79 &$\pm$&  9.19 \\ \hline
\multirow{2}{*}{CheXpert} 
& in $\mathcal{X}$ & 1.59 &$\pm$&  0.46 & 33.68 &$\pm$&  6.49 \\
& in $\mathcal{Z}$ & 13.69 &$\pm$&  4.54  & 34.71 &$\pm$&  6.51 \\ \hline
\multirow{2}{*}{Mall (min)} 
& in $\mathcal{X}$ & 1.34 &$\pm$&  0.39 & 19.11 &$\pm$&  2.67 \\
& in $\mathcal{Z}$ & 10.98 &$\pm$&  3.66  & 17.39 &$\pm$&  3.10 \\ \hline
\multirow{2}{*}{Mall (max)} 
& in $\mathcal{X}$ & 1.33 &$\pm$&  0.17 & 9.31 &$\pm$&  3.40 \\
& in $\mathcal{Z}$ & 15.90 &$\pm$&  5.28  & 19.35 &$\pm$&  6.36 \\ \hline
\end{tabular}
\caption{Euclidean norms $L^2$ in $\mathcal{X}$ for adversarial examples found via gradient ascent in $\mathcal{X}$ and counterfactuals found via gradient ascent in $\mathcal{Z}$. We show mean and standard deviation.}
\label{tab:L2_source_images_X}
\end{table}

\begin{table}[htbp]
\centering
\footnotesize
\begin{tabular}{|c|c|d{3.2}p{0.1cm}d{3.2}|d{3.2}p{0.1cm}d{3.2}|}
\hline
data set & img & \multicolumn{3}{c|}{$L^2$ source image}  & \multicolumn{3}{c|}{$L^2$ source class}  \\ \hhline{|=|=|===|===|}
\multirow{2}{*}{MNIST} 
& in $\mathcal{X}$ & 41.86 &$\pm$&  2.00 & 42.12 &$\pm$&  0.74 \\
& in $\mathcal{Z}$ & 35.26 &$\pm$&  4.68 & 39.91 &$\pm$&  1.61 \\ \hline
\multirow{2}{*}{CelebA} 
& in $\mathcal{X}$ & 473.72 &$\pm$&  171.49 & 542.21 &$\pm$&  149.63 \\
& in $\mathcal{Z}$ & 138.35 &$\pm$&  23.43 & 380.24 &$\pm$&  41.23 \\ \hline
\multirow{2}{*}{CheXpert} 
& in $\mathcal{X}$ & 355.50 &$\pm$&  114.23 & 539.18 &$\pm$&  88.40 \\
& in $\mathcal{Z}$ & 64.90 &$\pm$&  25.33 & 400.31 &$\pm$&  48.49 \\ \hline
\multirow{2}{*}{Mall (min)} 
& in $\mathcal{X}$ & 160.53 &$\pm$&  33.67 & 199.45 &$\pm$&  28.56 \\
& in $\mathcal{Z}$ & 78.56 &$\pm$&  11.84 & 180.39 &$\pm$&  15.78 \\ \hline
\multirow{2}{*}{Mall (max)} 
& in $\mathcal{X}$ & 142.50 &$\pm$&  7.61 & 153.96 &$\pm$&  8.80 \\
& in $\mathcal{Z}$ & 116.85 &$\pm$&  27.50 & 161.64 &$\pm$&  23.99 \\ \hline
\end{tabular}
\caption{Euclidean norms $L^2$ in $\mathcal{Z}$ for adversarial examples found via gradient ascent in $\mathcal{X}$ and counterfactuals found via gradient ascent in $\mathcal{Z}$. We show mean and standard deviation.}
\label{tab:L2_source_images_Z}
\end{table}

\subsection{Approximate Diffeomorphic Counterfactuals}\label{sec:experiments_approx}
In this section, we present our experimental analysis of approximate diffeomorphic counterfactuals for which we use variational autoencoders (VAEs)~\cite{kingma2013auto} and generative adversarial networks (GANs)~\cite{goodfellow2014generative}. As explained in Section~\ref{sec:approxDiffeo}, an important downside of approximate diffeomorphic counterfactuls is that the latent representation $z$ of the original image $x$ is generically lossy, i.e. $g(z) \neq x$, since the diffeomorphism is only approximate and not exact. On the other hand, an advantage of this approximation is that the method can be scaled to data of very high dimensionality. Both of these statements will be demonstrated experimentally in the following. 

\begin{figure*}[t]
\begin{center}
\centerline{\includegraphics[width=0.8\textwidth]{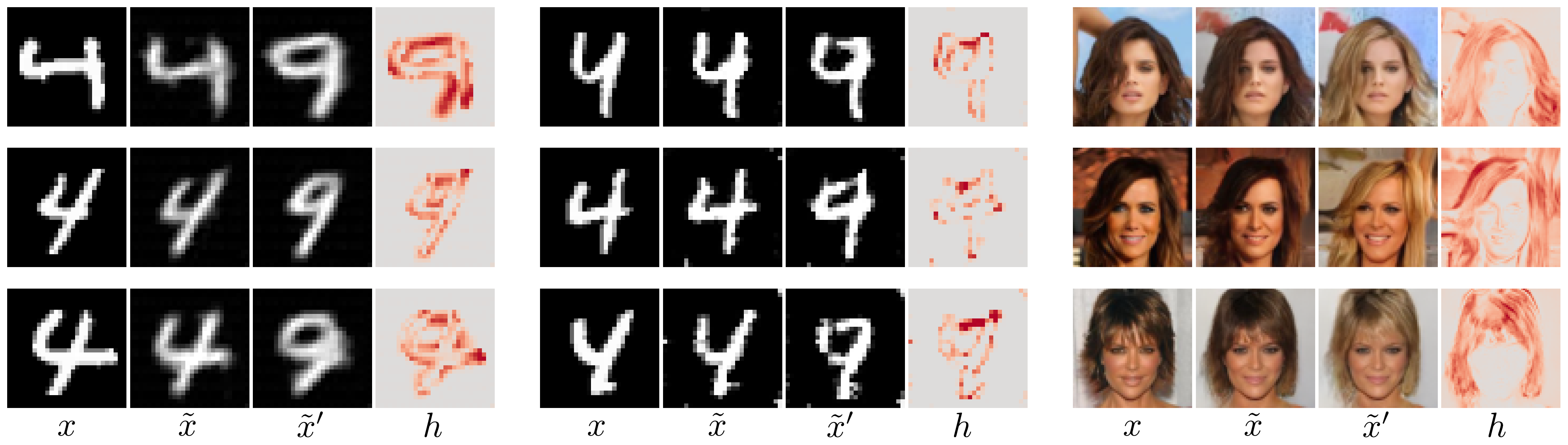}}
\caption{Counterfactuals for cVAE on MNIST (left block), dcGAN on MNIST (middle block) and pGAN on CelebA (right block). Columns of each block show original image, decoded latent representation of original, counterfactual and absolute difference $|\tilde{x}-\tilde{x}^\prime|$ summed over color channels.}
\label{fig:adv_attacks_gans_vae}
\end{center}
\vskip -0.2in
\end{figure*}

We use a simple convolutional VAE (cVAE) on the MNIST data set. Results are shown in Figure~\ref{fig:adv_attacks_gans_vae} in the left most block. The encoded and decoded image $\tilde{x}$ (second column of the block) appears slightly fuzzy but approximately reproduces the original image. Approximate diffeomorphic counterfactuals, found by gradient ascent in the latent space of the VAE, replicate features irrelevant for classification, such as stroke width and writing angle while structurally modifying the image so that it resembles an image of the digit nine.

As discussed in Section~\ref{sec:approxDiffeo}, GANs do generically not require an encoder during the training process and we apply GAN inversion methods to find an encoding of the source image. Specifically, for these relatively low-dimensional data sets, we find the latent representation $z$ by minimising the Euclidean norm between the decoded latent representation $g(z)$ and the original image $x$.
We apply our method to a simple convolutional GAN (dcGAN) for MNIST and a progressive GAN (pGAN)~\cite{karras2017progressive} for CelebA.  Results are shown in Figure~\ref{fig:adv_attacks_gans_vae} in the middle and right block. 

The dcGAN on MNIST produces some random pixel artifacts, but the generated images are sharper than those produced by the cVAE.

For the CelebA images generated with pGAN, we see that the decoded optimized latent representation of the original image deviates slightly from the original. This is especially visible if the composition is not typical (arm is not properly reproduced in the first row) or the background is highly structured (second row). For the approximate diffeomorphic counterfactuals, we observe even larger changes in the background. This may be attributed to the imperfect inversion process and the quality of the pGAN, i.e. the fact that the diffomorphism is only approximate and not exact.

\begin{figure}[t]
\begin{center}
\centerline{\includegraphics[width=1.0\columnwidth]{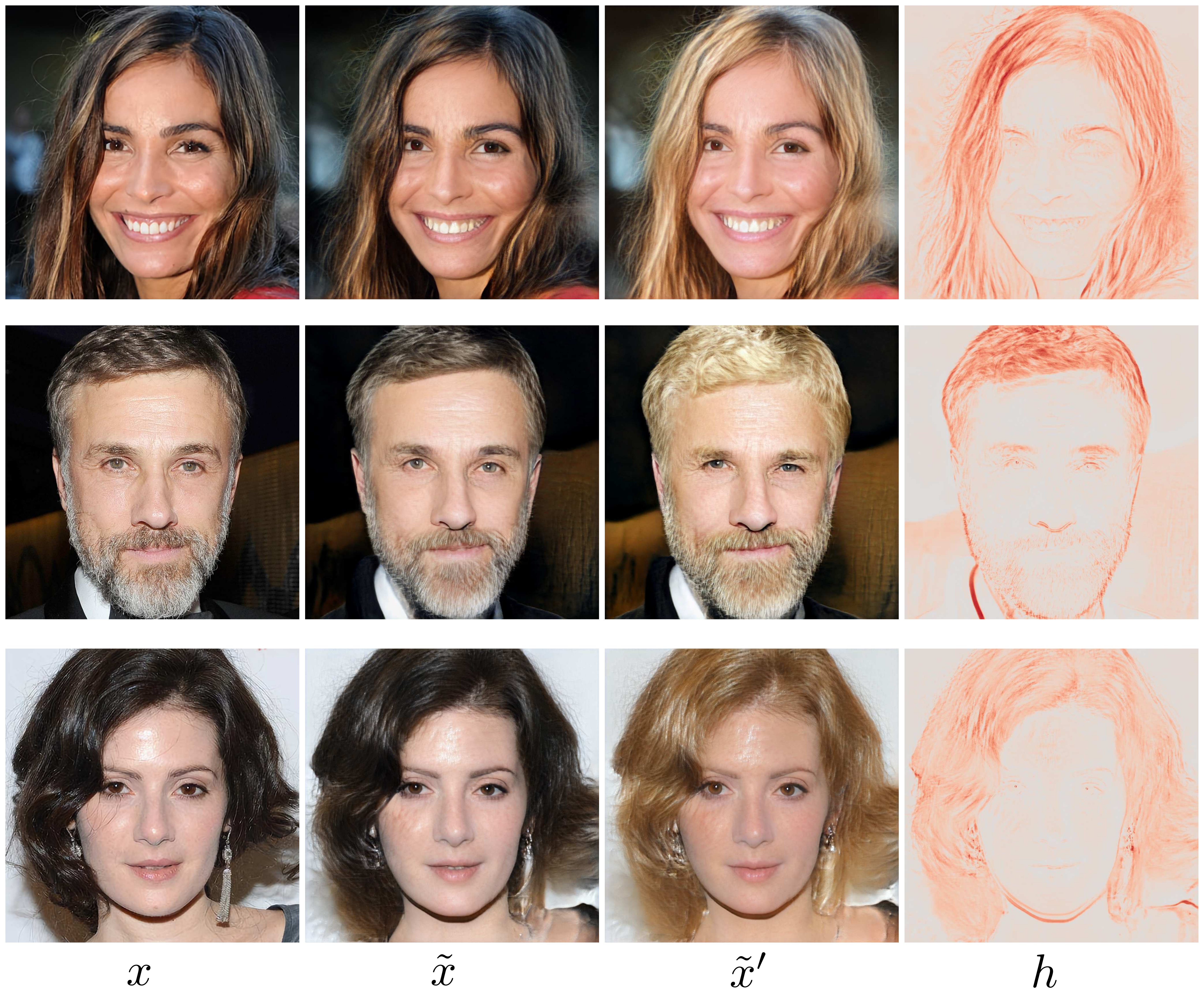}}
\caption{Counterfactuals generated with HyperStyle and Celeba-HQ. Columns show original, decoded latent representation, counterfactual and absolute difference $|\tilde{x}-\tilde{x}^\prime|$ summed over color channels.}
\label{fig:adv_attacks_hq}
\end{center}
\vskip -0.2in
\end{figure}

To demonstrate that approximate diffeomorphic explanations can scale to very high-dimensional data, we use a pretrained StyleGAN~\cite{karras2019style, karras2020analyzing} for images of resolution $1024\times1024$ from the CelebA-HQ data set~\cite{karras2017progressive}. To find the initial latent representation, we use HyperStyle~\cite{alaluf2021hyperstyle} GAN-inversion techniques. In order to use the same classifier as before, we downscale the images to $64\times64$ resolution before using them as input to the classifier. As demonstrated by Figure~\ref{fig:adv_attacks_hq}, approximate diffeomorphic counterfactuals lead to semantically meaningful and interpretable results even on these very high-dimensional data sets. 

\section{Related work}\label{sec:relatedworks}
In this section, we compare our approach with existing methods. 

This work builds upon and substantially extends our workshop contribution \cite{dombrowski2021diffeomorphic}. Specifically, we introduce both diffeomorphic and approximate diffeomorphic counterfactuals and discuss their relationship. To this end, we theoretically analyze this broader class of generative models in a unified manner which allows us to compare the relative strengths and weaknesses of these approaches. In addition, we provide an expanded discussion of the rigorous construction of the different coordinate systems. In our experimental studies, we consider additional generative models, specifically GANs and AEs, to find approximately diffeomorphic explanations for different data sets demonstrating that our method scales to a high-resolution data sets. Furthermore, we include experiments for additional datasets, such as CelebA-HQ and Mall, as well as further tasks, i.e. regression in addition to classification, and neural network architectures. We also consider a variety of quantitative evaluations of counterfactuals to evaluate the performance of the proposed methods.

\subsection{Counterfactuals with generative models}
A comparatively small number of publications consider normalizing flows, which started to gain attention relatively recently, in the context of generating counterfactuals. 

Hvilsh{\o}j et al.~\cite{hvilshoj2021ecinn} argue that a simple linear interpolation along the vector defined by the difference between two class centres in the base space of a flow can be interpreted as a counterfactual.
In contrast to our method, their approach does not involve the classifier and thus does not guarantee that the counterfactual is classified as the target class. In addition to that, their approach requires access to labelled training data, as the class specific centres in the base space have to be computed.

Sixt et al.~\cite{sixt2021interpretability} train a linear binary classifier directly in the base space of the flow. Adding the weight vector corresponding to the target class to the base space representation and projecting back to image space then produces a counterfactual with semantically changed features. Unlike our method, this approach requires training of a classifier and does not work for a general classifier of arbitrary architecture. In addition to that, their approach relies on the classifier being linear and binary so that the direction in which the image is modified can be determined analytically. Our method is more modular in the sense that the classifier can be pretrained and is independent of the generative model. Furthermore, we allow for the classifier to be non-binary as well as non-linear and we apply our framework for regression. 

Our approach is closest in spirit to the one taken by Joshi et al.~\cite{joshi2019towards}, who introduce an algorithm that does gradient ascent in the latent space of a generative model (they mention VAEs and GANs), while minimizing the difference between original and modified data point. The application concentrates on recourse for tabular data. The examples shown for image data are limited to a VAE, which results in relatively low quality counterfactuals, and lack quantitative evaluation.
Our approach focuses on image data and embeds their method into our more general framework with a detailed theoretical foundation. Furthermore, we propose to use normalizing flows as generative models for which the diffeomorphism is exact. We theoretically discuss the relation between diffeomorphic and approximately diffeomorphic methods and conduct extensive experiments for various tasks, data sets, and classifier architectures. To reliably evaluate the resulting counterfacuals we propose a number of approaches, specifically emphasising quantitative evaluation.

Other works also use autoencoders to generate counterfactuals.

Dhurandhar et al.~\cite{dhurandhar2018explanations} use elastic net regularization to keep the perturbation $\delta$ to the original data small and sparse. Furthermore, they use an autoencoder to minimize the reconstruction loss of the modified image and thus make sure the counterfactual lies on the data manifold. This approach was expanded by adding a prototype loss~\cite{looveren2021interpretable} that guides the counterfactuals towards the closest prototype that does not represent the source class and thus speeds up optimization and produces more interpretable counterfactuals. The prototypes are defined in the latent space of an autoencoder. Both approaches apply their algorithm on tabular data and MNIST.
Our approach differs from these works as we are not using generative models as a regularizer but directly modify the latent space representation. Our method also does not require access to labelled training data in order to compute prototypes, is applicable to high dimensional image data sets, and has no hyperparameters weighting different loss components.

Kim et al.~\cite{kim2021counterfactual} specifically train a Disentangled Causal Effect Variational Autoencoder (DCEVAE) and then generate counterfactuals conditioned on the original image and the label they aim to change. 
Our method on the other hand produces \emph{classifier dependent} counterfactuals and has thus potential to give insight into the decision processes of different classification models. Another advantage of our approach is that in contrast to~\cite{kim2021counterfactual}, the generative models for our method are not required to be trained with labeled training data.

A number of references use GANs to generate counterfactuals.

Zhao et al.~\cite{zhao2018generating} perturb the latent representation of a Wasserstein GAN using exhaustive search or continuous relaxation until they achieve a desired target classification. To get to the initial latent representation, they train an inverter. Zhao et al. refer to the resulting manipulated data points as \emph{natural adversarial examples} which we understand as effectively counterfactuals since they share characteristics like lying on the data manifold and having meaningful semantic deviations from the original. Although their counterfactuals in latent space are restricted to be close to the initial latent representation the resulting images deviate significantly from the original images for high dimensional data (LSUN), which makes it hard to identify features most relevant for the decision process. This could be due to inaccurate inversion or the randomized search process.
Our diffeomorphic counterfactuals can circumvent the problem of inaccurate inversion by using normalizing flows. In addition to that we use gradient ascent which has the advantage of potentially faster retrieval of counterfactuals and helps to guide changes to only relevant features. 

Chang et al.~\cite{chang2018explaining} use a conditional GAN for infilling regions that were previously removed from the original image. Their proposed algorithm aims to find an infilling mask which maximizes or minimizes the classification confidence while penalizing the size of the region that is replaced in the original. 
Compared to this, our approach has the advantage of not only directly influencing which pixels are changed but also how they are changed. We do not penalize the size of the region that changes as counterfactuals for different classes might have large differences in size of relevant features (for example a hair color change will require more changed pixels than a lipstick color change).

Samangouei et al.~\cite{samangouei2018explaingan} propose to use classifier prediction specific encoders together with a GAN which are trained to generate a reconstruction, a counterfactual, and a mask indicating which pixels should change between the counterfactual and the original. 
A similar approach is proposed by Singla et al.~\cite{singla2019explanation} as they also train a GAN, that is conditioned on the classifier predictions, jointly with an encoder to produce realistic looking counterfactuals. 
In both works, the classifier is incorporated in the training process of the GAN. After the training, the GAN generates counterfactuals without querying the classifier. As a consequence information about the classifier is integrated into the GAN purely during training, while our approach can be applied to independently trained models, which allows us to find counterfactuals for different classifiers using the same generative model.

Goetschalckx et al.~\cite{goetschalckx2019ganalyze} learn directions in the latent space by differentiating through a classifier and the generator so that cognitive properties of generated images, such as memorability, can be modified by moving in those directions. 
They do not specifically aim to produce counterfactuals but their approach touches on related concepts.
A difference to our work is that the latent representation is restricted to be modified along a single direction, while for our method the direction of change is dictated by the gradient over several update steps.

Lang et al.~\cite{lang2021explaining} train a StyleGAN with a classifier specific style space. The image can then be manipulated along the learned style coordinates. Our approach does not require training of additional models but can be applied to existing generators and classifiers. In contrast to our approach, their approach does not use a gradient-based approach to find directions that are sensitive to the classifier output. Instead, coordinate directions in style space that correspond to specific classes are detected by testing how changing a coordinate affects the classification of a number of samples. 

Lius et al.~\cite{liu2019generative} use a GAN specifically trained for editing that they condition on the original query image and the desired attributes. They apply gradient descent to find attributes that cause the GAN to generate an image that the classifier predicts as the target class, while at the same time enforcing the image to be close to the original. In contrast to this approach, we directly modify the latent representation rendering our approach independent of the exact structure of the generative model. We also observe that our counterfactuals stay close to the original image without explicitly enforcing similarity.

Shen et al.~\cite{shen2020interfacegan} train a linear SVM in the latent space of a GAN using generated images, for which the facial attributes are determined after projection to the image space using a pretrained classifier. They can then modify the latent representation of an inverted image linearly along the normal directions of the learned SVMs. In contrast to our approach their method requires sampling and training. They do not use gradients and can only linearly modify the latent code. The classifier is only indirectly used for labeling the generated samples prior to training the SVM.

\subsection{Quantitative Metrics for Counterfactuals}
Many of the above works are limited to qualitative assessment of counterfactuals. The quantitative assessment of counterfactuals is still an active research area, a summary of quantitative measures can be found in~\cite{hvilshoj2021quantitative}. 

The most reliable evaluation of counterfactuals can be obtained by a large study that lets human agents evaluate the generated counterfactuals. As those are relatively costly to conduct and can introduce unexpected biases if not designed carefully, their application is often infeasible. Nevertheless, a few works~\cite{goyal2019counterfactual, zhao2018generating, sixt2021interpretability, lang2021explaining} undertake small user studies ($9\leq N \leq 60$) on a relatively limited set of generated counterfactuals. We aim to approximate an independent human evaluation by testing our counterfactuals on newly trained models that serve as oracles.

Some works\cite{rombach2020making, kim2021counterfactual, singla2019explanation} apply a metric commonly used for generative model evaluation, the Fréchet Inception Distance (FID) score~\cite{heusel2017gansFID}, measuring the quality of the generated explanation compared to samples from the data set. As we find counterfactuals by moving directly in latent space the FID for our counterfactuals would be very similar to the FID of the generative model itself. We therefore do not consider the FID to be a meaningful metric for (approximate) diffeomorphic counterfactuals.

Van Looveren and Klaise~\cite{looveren2021interpretable} propose two metrics to test interpretability: IM1 (defined in \eqref{eq:12} above) uses two autoencoders which were each trained on data from only one class and computes the relative reconstruction error of the counterfactual. The second metric IM2 (defined in \eqref{eq:13} above) calculates the normalized difference between a reconstruction of an autoencoder trained on the target class and an autoencoder trained on all classes.
Van Looveren and Klaise use these two metrics to compare how different loss functions effect the relative interpretability measured by IM1 and IM2 for the MNIST data set. We limit the quantitative evaluation of our counterfactuals to IM1, since IM2 has been subject to controversies.

Other works~\cite{samangouei2018explaingan, kim2021counterfactual} check substitutability. They train classifiers on a training set consisting of generated counterfactuals and compare their performance on the original test data set to a classifier trained on the original training set. As we can generate counterfactuals that are classified with very different confidence, this method may not be useful as results may be highly dependent on the choice of confidence.

Other methods aim to evaluate explanations by replacing pixel values or entire regions based on the importance of features in the explanation~\cite{bach2015pixelwise, samek2017evaluating, arras2017explainingRNN, tomsett2020sanityChecksMetrics, samangouei2018explaingan} and testing the performance of a classifier on the modified images. Those methods may suffer from creating images that lie off the data manifold, so that a thorough comparison may require extensive retraining~\cite{hooker2019roar}.

\section{Conclusion}
In this work, we proposed theoretically rigorous yet practical methods to generate counterfactuals for both classification as well as regression tasks, namely exact and approximate diffeomorphic counterfactuals. The exact diffeomorphic counterfactuals are obtained by following gradient ascent in the base space of a normalizing flow. While approximate diffeomorphism are obtained with the help of either generative adversarial networks or variational autoencoders. Our thorough theoretical analysis, using Riemannian differential geometry, shows that for well-trained models, our counterfactuals necessarily stay on the data manifold during the search process and consequently exhibit semantic features corresponding to the target class. Approximate diffeomorphic counterfactuals come with the risk of information loss but allow excellent scalability to higher dimensional data.
Our theoretical findings are backed by experiments which both quantitatively and qualitatively demonstrate the performance of our method on different classification as well as regression tasks and for numerous data sets.

The application of our counterfactual explanation method is straightforward and requires no retraining, so that it can be readily applied to investigate common problems in deep learning like identifying biases for classifiers or training data or scrutinizing falsely classified examples --- all common tasks for applications in computer vision and beyond. 

For future work, we intend to investigate the benefit of our counterfactuals in the sciences, in particular for medical applications such as digital pathology~\cite{binder2021morphological} or brain computer-interfaces~\cite{lemm2011introduction}.

Furthermore, the method presented in this work is not restricted to the generation of counterfactuals for image data or in computer vision. In particular, one could imagine applications e.g.\ in chemistry and physics where the technique proposed here may be used to optimize desired properties of stable molecules which are restricted to minima of the associated potential energy surface \cite{unke2021, gebauer2019symmetry, gebauer2022inverse, garcia2021n}. 

In practical applications, it is often beneficial to incorporate symmetries as inductive bias following a well-established paradigm in machine learning \cite{chmiela2018towards, chmiela2017machine}. It is straightforward to incorporate symmetries into our method by employing equivariant normalizing flows as constructed e.g.\ in \cite{kohler2020equivariant}.

In conclusion, our method is applicable to a broad range of computer vision problems and beyond as it provides a way to optimize the output of a predictive model on the data manifold given only indirectly by a trained generative model.





%

\section*{Acknowledgment}

This work is supported by the Berlin Institute for the Foundations of Learning and Data (BIFOLD). KRM was partly supported by the Institute of Information \& Communications Technology Planning \& Evaluation (IITP) grants funded by the Korea government(MSIT) (No. 2019-0-00079, Artificial Intelligence Graduate School Program, Korea University and No. 2022-0-00984, Development of Artificial Intelligence Technology for Personalized Plug-and-Play Explanation and Verification of Explanation). This work was supported by the German Ministry for Education and Research (BMBF) under Grants 01IS14013A-E, 01GQ1115, 1GQ0850, 01IS18025A and 01IS18037A. JG is supported by the Swedish Research Council and by the Knut and Alice Wallenberg Foundation. PK also wants to thank Shinichi Nakajima and Maximilian Alber for insightful discussions. Correspondence to PK and KRM.

\bibliographystyle{unsrt}

\bibliography{main.bib}

\clearpage
\appendices

\section{Proofs}
\subsection{Proof of Theorem~\ref{th:diff_ctfctls}}\label{app:diff_ctfctls_proof}
In this section, we provide the proof for Theorem~\ref{th:diff_ctfctls}, which we repeat here for completeness.
\setcounter{theorem}{0}
\begin{theorem}
    For $\epsilon\in(0,1)$ and $g$ a normalizing flow with Kullback--Leibler divergence $\KL(p,q)<\epsilon$,
    \begin{align*}
        \gamma_{\perp_i}^{-1}\rightarrow0\qquad\text{as}\qquad\delta\rightarrow0
    \end{align*}
    for all $i\in \{1,\dots,N_\mathcal{X}-N_\mathcal{D} \}$.
\end{theorem}
Since normalizing flows are diffeomorphisms, $g^{-1}$ exists and is differentiable, $\mathcal{Z}=\mathcal{X}$ and $\gamma$ is a non-singular metric on all of $\mathcal{X}$. Furthermore, the base distribution $q:\mathcal{X}\rightarrow\mathbb{R}$ transforms like a density,
\begin{align}
  q_{x}(x)=q_{z}(g^{-1}(x))\left|\frac{\partial z^{a}}{\partial x^{\alpha}}\right|\,,\label{eq:10}
\end{align}
where $q_{x,z}$ denote $q$ in $z^{a}$ and $x^{\alpha}$ coordinates, respectively,  $q_{x,z}:\mathbb{R}^{N_{\mathcal{X}}}\rightarrow\mathbb{R}$. We will assume that $q_{z}$ is the univariate Gaussian distribution.

We assume that the Kullback--Leibler divergence between $p$ and $q$ is small, i.e.\ that
\begin{align}
  \KL(p,q)<\epsilon
\end{align}
for some small $\epsilon\in(0,1)$. Then, since $\ln(1/a)\geq 1-a$,
\begin{align}
  \epsilon &> \int_{S_{x}}p_{x}(x)\ln\left( \frac{p_{x}(x)}{q_{x}(x)} \right)\dd{x}\nonumber\\
  &\geq \int_{S_{x}}p_{x}(x)\left( 1- \frac{q_{x}(x)}{p_{x}(x)} \right)\dd{x}\nonumber\\
  &=1-\int_{S_{x}}q_{x}(x)\dd{x}
\end{align}
and therefore
\begin{align}
  \int_{S_{x}}q_{x}(x)\dd{x}>1-\epsilon\,.\label{eq:11}
\end{align}
Intuitively, this means that most of the induced probability mass lies in the support of $p$.

We now write $q$ in $z^{a}$ coordinates using \eqref{eq:10} and then evaluate the integral in $y^{\mu}$ coordinates,
\begin{align}
  1-\epsilon &< \int_{S_{x}}q_{z}(g^{-1}(x))\left| \frac{\partial z^{a}}{\partial x^{\alpha}} \right|\dd x\nonumber\\
  &= \int_{S_{y}}q_{z}(g^{-1}(x(y)))\left| \frac{\partial z^{a}}{\partial x^{\alpha}} \right|\,\left| \frac{\partial x^{\alpha}}{\partial y^{\mu}} \right|\dd{y}\,.
\end{align}
Using the block-diagonal form \eqref{eq:7} of the metric in $y^{\mu}$ coordinates, the integration measure simplifies to
\begin{align}
  \left| \frac{\partial z^{a}}{\partial x^{\alpha}} \right|\,\left| \frac{\partial x^{\alpha}}{\partial y^{\mu}} \right| = \sqrt{|\gamma_{\mu\nu}|}=\sqrt{|\gamma_{\mathcal{D}}|}\ \prod_{i=1}^{N_{\mathcal{X}}-N_{\mathcal{D}}}\ \sqrt{|\gamma_{\perp_{i}}|}\,.
\end{align}
Therefore, we have
\begin{align}
  1-\epsilon < \int_{\mathcal{D}_{y}}\sqrt{|\gamma_{\mathcal{D}}|}\ \prod_{i=1}^{N_{\mathcal{X}}-N_{\mathcal{D}}}\ \int_{-\delta/2}^{\delta/2}\sqrt{|\gamma_{\perp_{i}}|}q_{z}(z(y))\dd{y_{\perp}^{i}}\dd{y_{\parallel}}\,.
\end{align}
Since $q_{z}$ is bounded, as $\delta\rightarrow0$, we need $|\gamma_{\perp_{i}}|\rightarrow\infty$ in order to keep the integral above the bound. Therefore, $\gamma_{\perp_{i}}^{-1}\rightarrow0$ for $\delta\rightarrow0$.

\subsection{Proof of Theorem~\ref{th:approx_diff_ctfctls}}\label{app:proof_approx_diff_ctfctls_proof}
In this section, we provide the proof for Theorem~\ref{th:approx_diff_ctfctls}, which we repeat here for convenience. 
\begin{theorem}\label{th:approx_diff_ctfctls}
    If $g:\mathcal{Z}\rightarrow\mathcal{X}$ is a generative model with $\mathcal{D}\subset g(\mathcal{Z})$ and image $g(\mathcal{Z})$ which extends in any non-singular orthogonal direction $y_{\perp}^{i}$ outside of $\mathcal{D}$,
    \begin{align*}
        \gamma_{\perp_{i}}^{-1}\rightarrow 0
    \end{align*}
    for $\delta\rightarrow0$ for all non-singular orthogonal directions $y_{\perp}^{i}$.
\end{theorem}
For any $x_{\mathcal{D}}\in\mathcal{D}$, let $x_{0}\in S$ be on the negative $y_{\perp}^{i}(x_{\mathcal{D}})$ coordinate line such that $p(x_{0})<\epsilon$ for some small $\epsilon$ and let $x_{1}\in S$ be on the positive $y_{\perp}^{i}(x_{\mathcal{D}})$ coordinate line such that $p(x_{1})<\epsilon$, as illustrated in Figure~\ref{fig:tau}. Then, the assumption that $g(\mathcal{Z})$ extends beyond $\mathcal{D}$ in non-singular directions implies that the segment of the $y_{\perp}^{i}(x_{\mathcal{D}})$ coordinate line between $x_{0}$ and $x_{1}$ lies entirely in $g(\mathcal{Z})$.

Let $\tau:[0,1]\rightarrow S$ be the coordinate-line segment between $x_{0}$ and $x_{1}$. In summary, we have
\begin{enumerate}
\item \label{item:3}$\tau(0)=x_{0}$ and $\tau(1)=x_{1}$ with $p(x_{0})<\epsilon$ and $p(x_{1})<\epsilon$
\item \label{item:2}$\exists$ $t_{\mathcal{D}}\in[0,1]$ such that $\tau(t_{\mathcal{D}})=x_{\mathcal{D}}\in\mathcal{D}$
\item \label{item:1}$\tau_{\perp}^{j}=0$ for $j\neq i$, $\tau_{\parallel}=\mathrm{const.}$
\end{enumerate}

In particular, \ref{item:1}) implies that the tangent vector of $\tau$ points along the $y_{\perp}^{i}$ coordinate vector: $\tau'(t)\propto\partial_{y_{\perp}^{i}}$.

Let $\mathcal{L}(\tau)$ denote the length of $\tau$, i.e.
\begin{align}
  \mathcal{L}(\tau)&=\int_{0}^{1}\sqrt{\gamma(\tau'(t),\tau'(t))}\,\dd{t}\\
  &=\int_{0}^{1}\sqrt{\gamma_{\mu\nu}(\tau(t))\frac{\dd{\tau^{\mu}}}{\dd{t}}\frac{\dd{\tau^{\nu}}}{\dd{t}}}\,\dd{t}\,.
\end{align}
Following point \ref{item:1}) above, we can perform the implicit sums over $\mu$ and $\nu$ and get
\begin{align}
  \mathcal{L}(\tau)&=\int_{0}^{1}\sqrt{\gamma_{\perp_{i}}(\tau(t))}\,\Big|\frac{\dd{\tau^{i}}}{\dd{t}}\Big|\,\dd{t}\\
  &=\int_{x_{0,\perp}{}^{i}}^{x_{1,\perp}{}^{i}}\sqrt{\gamma_{\perp_{i}}(y_{\perp}^{i})}\,\dd{y_{\perp}^{i}}\,.
\end{align}
Since $S$ has, by construction, (Euclidean) extension $\delta$ orthogonal to $\mathcal{D}$ in $y^{\mu}$ coordinates, with $\delta\ll 1$, we perform a Taylor expansion of the metric around $x_{\mathcal{D}}$
\begin{align}
  \gamma_{\mu\nu}(\tau(t))&= \gamma_{\mu\nu}(x_{\mathcal{D}})+\mathcal{O}(\tau^{\rho}(t)-x_{\mathcal{D}}^{\rho})
\end{align}
and obtain to first order
\begin{align}
  \mathcal{L}(\tau)\approx\sqrt{\gamma_{\perp_{i}}(x_{\mathcal{D}})}\,(x_{1,\perp}{}^{i}-x_{0,\perp}{}^{i})\,.
\end{align}
Again, since $S$ has range $\delta$ in $y_{\perp}^{i}$-direction, we have $(x_{1,\perp}{}^{i}-x_{0,\perp}{}^{i})<\delta$ and therefore
\begin{align}
  \gamma_{\perp_{i}}>\frac{\mathcal{L}^{2}(\tau)}{\delta^{2}}\,.\label{eq:8}
\end{align}


We now change the $x^{\alpha}$- and $y^{\mu}$ coordinates such that $\delta\rightarrow 0$, corresponding to a data distribution which is more and more concentrated on $\mathcal{D}$. As we change coordinates, $\mathcal{L}(\tau)$ is constant as a geometric invariant\footnote{Since $\tau$ lies entirely in $g(\mathcal{Z})$, there is a curve $\sigma$ in $\mathcal{Z}$ whose image under $g$ is $\tau$. Together with the properties \ref{item:3}) and \ref{item:2}) of $\tau$ this implies in particular that $\mathcal{L}(\tau)=\mathcal{L}(\sigma)$ is not infinitesimal.} and we obtain from~\eqref{eq:8}
\begin{align}l
    \gamma_{\perp_{i}}^{-1}\rightarrow 0\,,\label{eq:9}
\end{align}
as desired.

\section{Details on Experiments}\label{app:experiments}

\subsection{Toy Example}
\label{sec:toy_sample}
The flow used for the toy example is composed of 12 RealNVP-type coupling layer blocks. Each of these blocks includes a three-layer fully-connected neural network with leaky ReLU activations for the scale and translation functions.

For training, we sample from the target distribution defined by

\begin{align*}
    x_3 &\sim U(-4, 4) \,,\\
    x_2 &= \cos(x_3) \,,\\
    x_1 &= \sin(x_3) \,.
\end{align*}

\begin{figure}[t]
\centering
\includegraphics[width=1.0\linewidth]{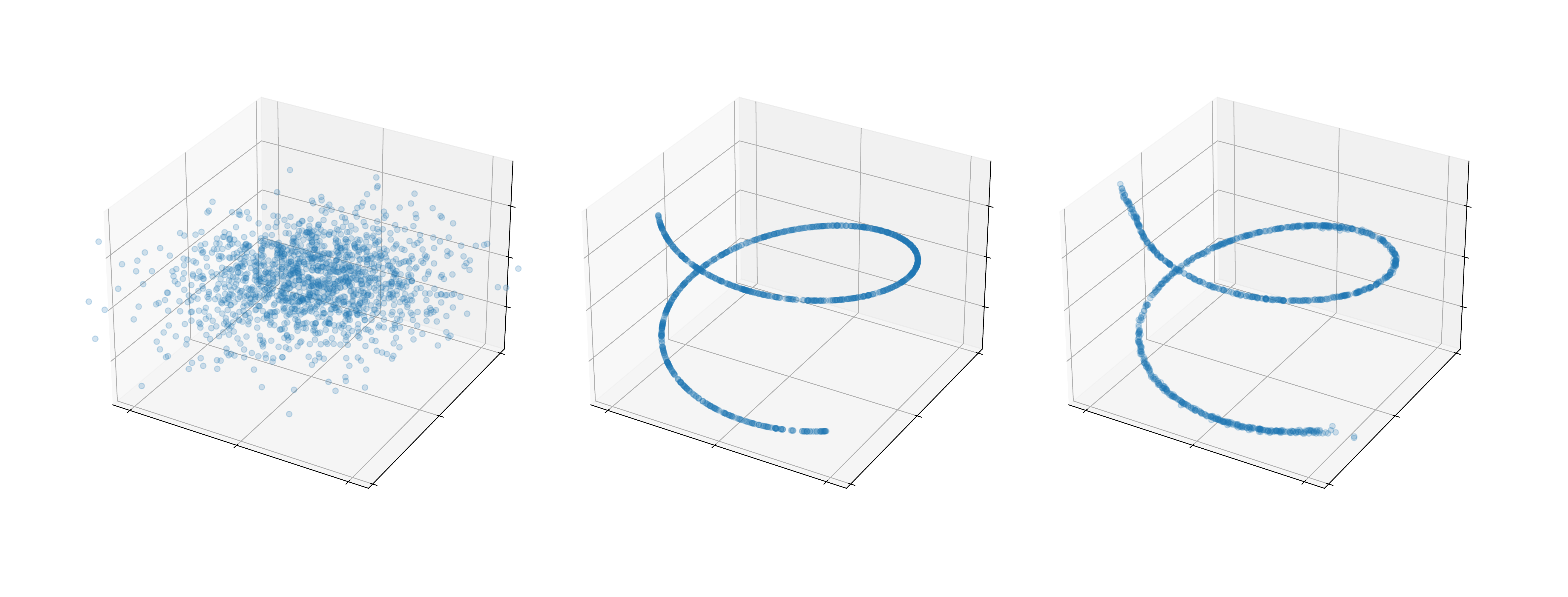}
\caption{From left to right: distribution in the base space of the flow, target distribution, learned distribution}
\label{fig:toy_distribution}
\end{figure}

We train for 5000 epochs using a batch of 500 samples per epoch. We use the Adam optimizer with standard parameters and learning rate $\lambda = 1\times 10^{-4}$. This takes around 10 minutes on a standard CPU. 
After successful training we can map samples from a multivariate standard Normal distribution to the data distribution, see Figure~\ref{fig:toy_distribution}.

In order to train a classifier we first define the ground truth: points with z-coordinate smaller than zero belong to the one class and points with z-coordinate bigger than zero belong to the other class. 
We train a neural network with 256 hidden neurons with ReLU activations and one output neuron with sigmoid activation to near perfect accuracy on this classification task. 

We then run the gradient ascent optimization in image space $\mathcal{X}$ and in the base space of the flow $\mathcal{Z}$. We start from samples from the true data distribution and set the target to 0.1 if the network predicted a value larger than 0.5 for the original data point, otherwise we set the target to 0.9.

\begin{figure}[t]
\centering
\includegraphics[width=0.6\linewidth]{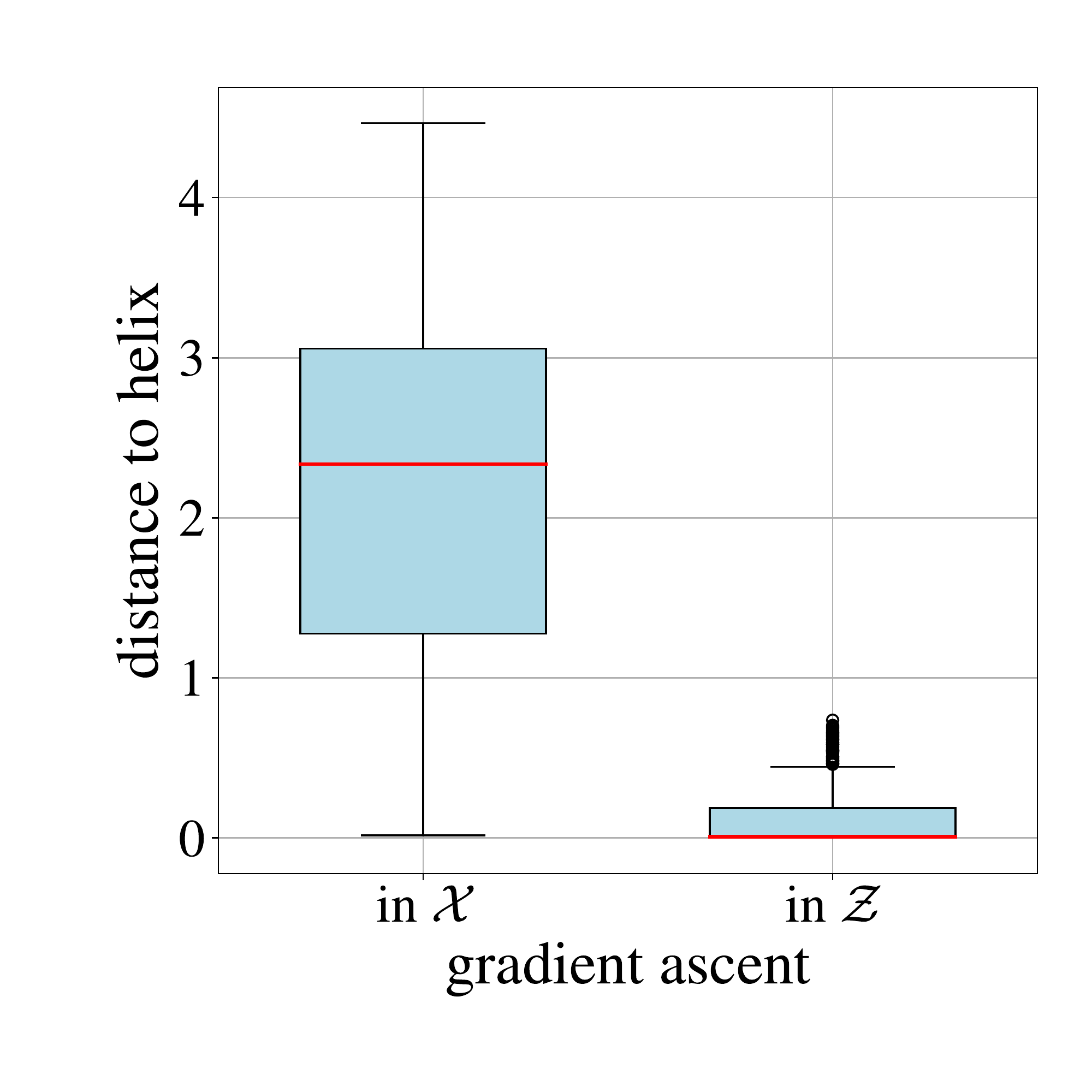}
\caption{Statistics for 1000 counterfactuals/adversarials. Boxes extend from lower to upper quartile, red lines mark the medians, whiskers mark the 1.5$\times$IQR and circles mark outliers.}
\label{fig:toy_quantitative}
\end{figure}

Figure~\ref{fig:toy_quantitative} shows that counterfactuals found in $\mathcal{Z}$ lie significantly closer to the data manifold than adversarials found in $\mathcal{X}$

For more details we refer to our github implementation\footnote{\url{https://github.com/annahdo/counterfactuals/blob/main/toy_example.ipynb}}.

\subsection{Data sets:}
We test our method on four different data sets.

\underline{MNIST:} We use training and test data as specified in torchvision. We use 10\% of the training data for validation.

\underline{CelebA:} We take the training and test data set as specified in torchvision. We use 10\% of the training images for validation.
We scale and center crop the images to 64$\times$64 pixels. We augment the training data by horizontally flipping the images.

\underline{CheXpert:} We choose the first 6500 patients from the training set for testing. The remaining patients are used for training. We select the model based on performance on the original validation set.
We only consider frontal images and scale and crop the images to 128$\times$128 pixels.

\underline{Mall:}
We resize the images so that the shortest side has 128 pixels. We then take a $64\times 64$ pixel cutout (starting from pixel $[r=64, c=100]$)
We use 1620 images for training, 180 images for evaluation and 200 images for testing. 

\subsection{Flows}
\begin{figure*}[!t]
\captionsetup[subfigure]{labelformat=empty}
\centering
\subfloat[(a)]{
\includegraphics[width=.49\columnwidth]{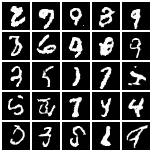}}\hfill
\subfloat[(b)]{\includegraphics[width=.49\columnwidth]{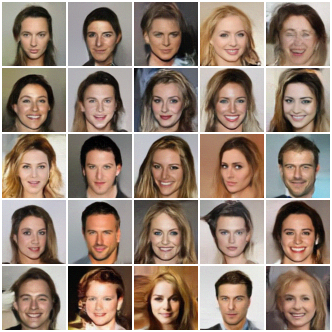}}\hfill
\subfloat[(c)]{\includegraphics[width=.49\columnwidth]{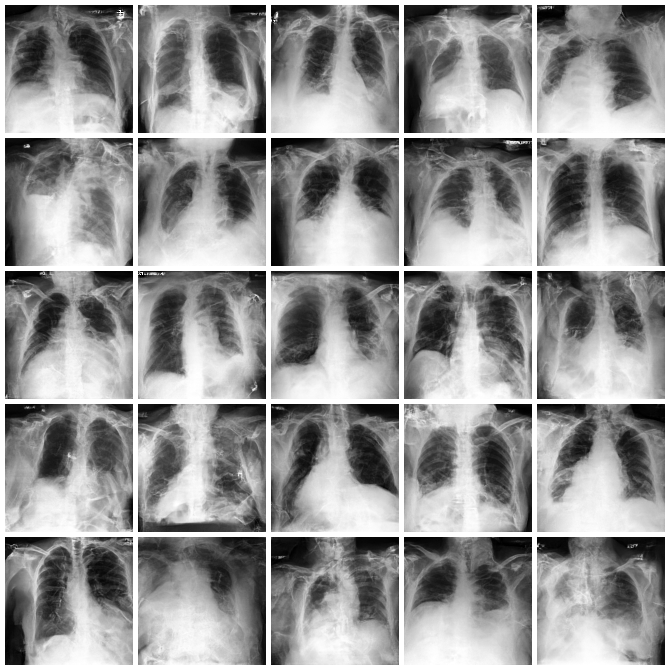}}\hfill
\subfloat[(d)]{\includegraphics[width=.49\columnwidth]{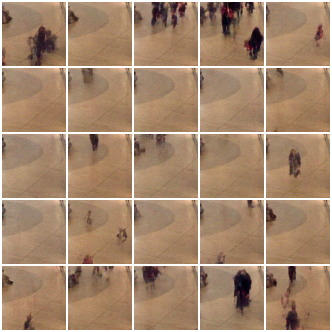}}
\label{fig:samples_flows}
\caption{Generated samples for all normalizing flows used in the paper: RealNVP on MNIST (a), Glow on CelebA (b), Glow on CheXpert (c) and Glow on Mall (d).}
\end{figure*}

We show generated samples for all Flows in Figure~\ref{fig:samples_flows}.

\textbf{Architecture:} We use the RealNVP~\footnote{ \url{https://github.com/fmu2/realNVP}} architecture for MNIST and the Glow~\footnote{ \url{https://github.com/rosinality/glow-pytorch}} architecture for CelebA and CheXpert. 

\textbf{Training:}
We use the Adam optimizer with a learning rate of $1\times 10^{-4}$ and weight decay of $5\times 10^{-4}$ for all flows.

\underline{MNIST:} we train for 30 epochs on all available training images. Bits per dimension on the test set average to 1.21.

\underline{CelebA:} we train for 8 epochs on all available training images. We use 5 bit images. Bits per dimension on the test set average to 1.32.

\underline{CheXpert:} we train for 4 epochs on all available training images. Bits per dimension on the test set average to 3.59.

\underline{Mall:} we train for 47 epochs on all available training images. Bits per dimension on the test set average to 0.96.

\subsection{Classifier}
\textbf{Architecture:} All classifiers have a similar structure consisting of convolutional, pooling and fully connected layers. We use ReLU activations and batch normalization. For MNIST we use four convolutional layers and three fully connected layers. For CelebA and CheXpert we use six convolutional layers and four fully connected layers.

\textbf{Training:} We use the Adam optimizer with a weight decay of $5\times 10^{-4}$ for all classifiers.

\underline{MNIST:} We train for 4 epochs using a learning rate of $1\times 10^{-3}$. We get a test accuracy of 0.99.

\underline{CelebA:} We train a binary classifier on the blond attribute.
We partition the data sets into all images for which the blond attribute is positive and the rest of the images. We treat the imbalance by undersampling the class with more samples. We train for 10 epochs using a learning rate of $5\times 10^{-3}$. We get a balanced test accuracy of 93.63\% by averaging over true positive rate (93.95\%) and true negative rate (93.31\%).

\underline{CheXpert:} We train a binary classifier on the cardiomegaly attribute.
For the training data the cardiomegaly attibute can have four different values: blanks, 0, 1, and -1. We label images with the blank attribute as 0 if the no finding attribute is 1, otherwise we ignore images with blank attributes. We also ignore images where the cardiomegaly attribute is labeled as uncertain. Using this technique, we obtain 25717 training images labelled as healthy and 20603 training images labelled as cardiomegaly. We do not treat the imbalance but train on the data as is. We train for 9 epochs using a learning rate of $1\times 10^{-4}$. We test on the test set, that was produced in the same way as the training set. We get a balanced test accuracy of 86.07\% by averaging over true positive rate (84.83\%) and true negative rate (87.27\%). 

\subsection{U-Net:}
The U-Net~\cite{ronneberger2015unet} follows an hour glass structure. The first part consists of multiple convolutional, batch normalization, ReLU and pooling layers that gradually reduce the spatial dimensions while increasing the channel dimensions. The second block consists of upsampling, concatenation of feature maps from the first part, convolutional, batch normalization and ReLU activation layers. The last layer has the same spacial dimension as the input but only one channel corresponding to a probability map. For using the U-Net in order to count pedestrians, Ribera et. al.~\cite{ribera2019locating} add an additional fully connected layer with ReLU activations that combines the information from the last layer and the central encoding layer to estimate the number of objects of interest present~\footnote{ \url{https://github.com/javiribera/locating-objects-without-bboxes}}.

\subsection{Optimization of counterfactuals and adversarial examples}
\begin{figure}[t]
\vskip 0.2in
\begin{center}
\centerline{\includegraphics[width=.99\columnwidth]{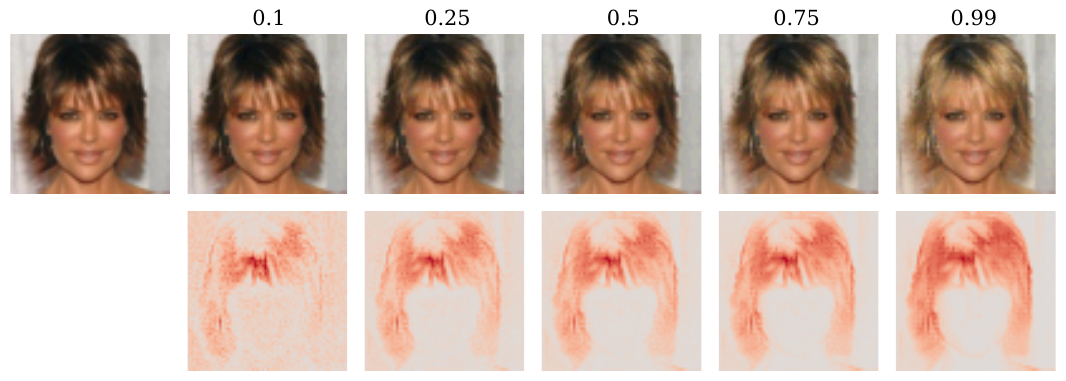}}
\caption{Top row: original image and evolution throughout optimization. Numbers indicate confidence with which the image is classified as `blond'. Second row: absolute differences to original image summed over color channels.}
\label{fig:adv_attack_evolution}
\end{center}
\vskip -0.2in
\end{figure}

Counterfactuals and adversarial examples are found using the Adam optimizer with standard parameters. We vary only the learning rate $\lambda$. For our main experiments we use the base space of normalizing flows to find the counterfactuals.
We set the threshold for the confidence of the target class high when searching for counterfactuals and adversarial examples. We therefore get more visually expressive results. Of course in practice one might whish to find counterfactuals with lower target confidence. We show an example optimization with different confidence thresholds in Figure~\ref{fig:adv_attack_evolution}.

\underline{MNIST:} We use $\lambda = 5\times 10^{-4}$ for conventional adversarial examples and $\lambda = 5\times 10^{-2}$ for counterfactuals found via the flow. We do a maximum of 2000 steps stopping early when we reach the target confidence of 0.99. We perform attacks on 500 images of the true class `four'. All conventional attacks and 498 of the attacks via the flow reached the target confidence of 0.99 for the target class `nine'. 

\underline{CelebA:} We use $\lambda = 7\times 10^{-4}$ for conventional adversarial examples and $\lambda = 5\times 10^{-3}$ for counterfactuals found via the flow. We do a maximum of 1000 steps stopping early when we reach the target confidence of 0.99. We perform attacks on 500 images of the true class `not-blond'. 492 conventional attacks and 496 of the attacks via the flow reached the target confidence of 0.99 for the target class `blond'.

\underline{CheXpert:} We use $\lambda = 5\times 10^{-4}$ for conventional adversarial examples and $\lambda =5\times 10^{-3}$ for counterfactuals found via the flow. We do a maximum of 1000 steps stopping early when we reach the target confidence of 0.99. We perform attacks on 1000 images of the true class `healthy'. All conventional attacks and 990 of the attacks via the flow reached the target confidence of 0.99 for the target class `cardiomegaly'. 

\underline{Mall:} We use $\lambda = 1\times 10^{-4}$ for conventional adversarial examples and $\lambda =5\times 10^{-3}$ for counterfactuals found via the flow. We do a maximum of 5000 steps stopping early when we reach the target regression value of 10 when we are maximizing pedestrians and 0.01 when we are minimizing pedestrians. We perform attacks on 100 images with few people (average regression value of 0.7) and 100 images with many people (average regression value of 3.6). All attacks reached the target values.

\underline{GANs:}
For counterfactuals found in the latent space of GANs we do a maximum of 1000 steps with $\lambda=5\times 10^{-3}$ for MNIST, CelebA and CelebA-HQ. 

\underline{VAEs:}
For counterfactuals found in the latent space of the VAE trained on MNIST we do a maximum of 1000 steps with $\lambda=5\times 10^{-3}$.

\subsection{Similarity to source images}\label{app:experiments_L2_source_images}

\begin{figure}[t]
\begin{center}
\centerline{\includegraphics[width=1.0\columnwidth]{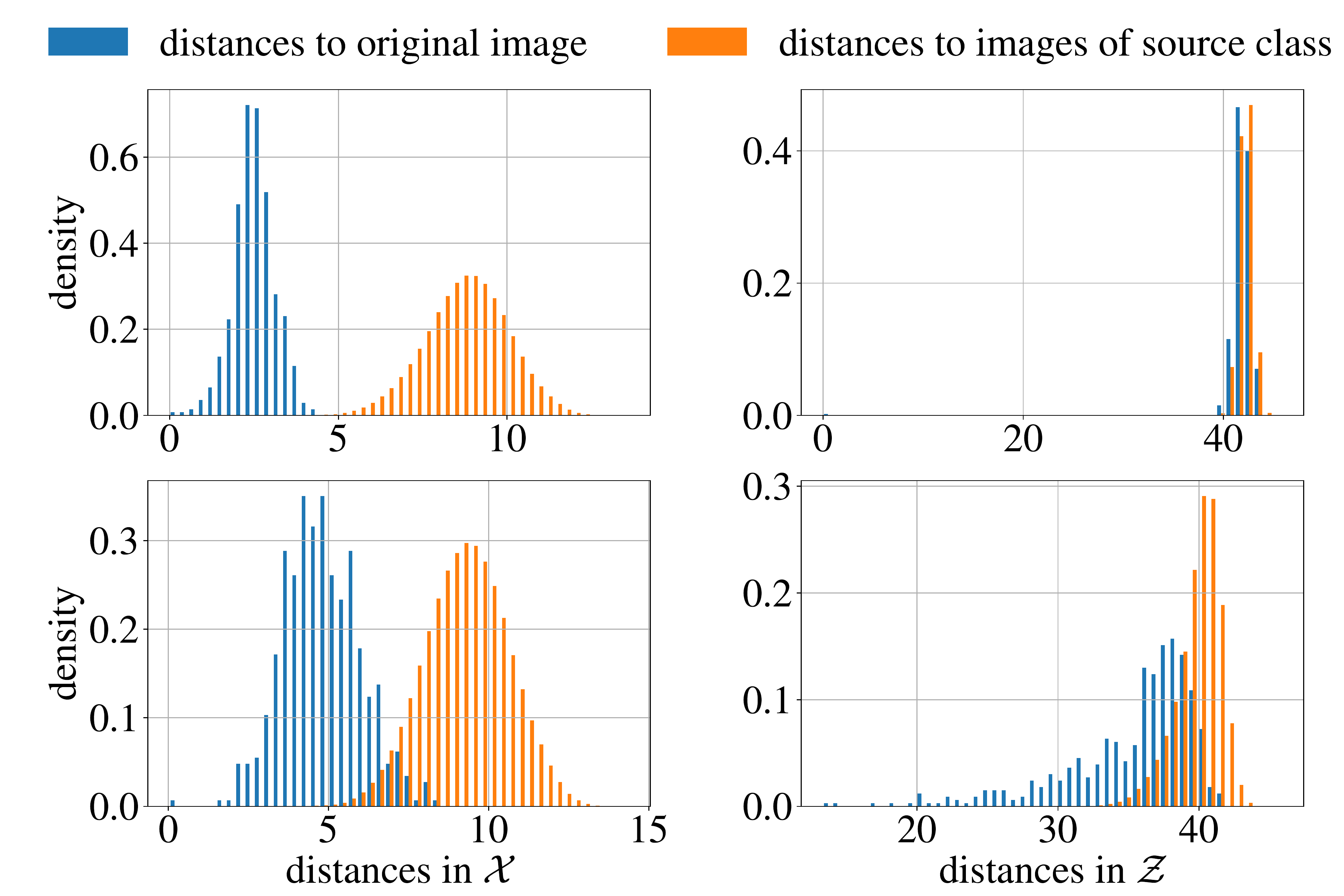}}
\caption{Euclidean distances in $\mathcal{X}$ and $\mathcal{Z}$ for adversarial examples (top row) and counterfactuals (bottom row) for the MNIST data set.}
\label{fig:L2_source_MNIST_conv_z}
\end{center}
\vskip -0.2in
\end{figure}

\begin{figure}[t]
\begin{center}
\centerline{\includegraphics[width=1.0\columnwidth]{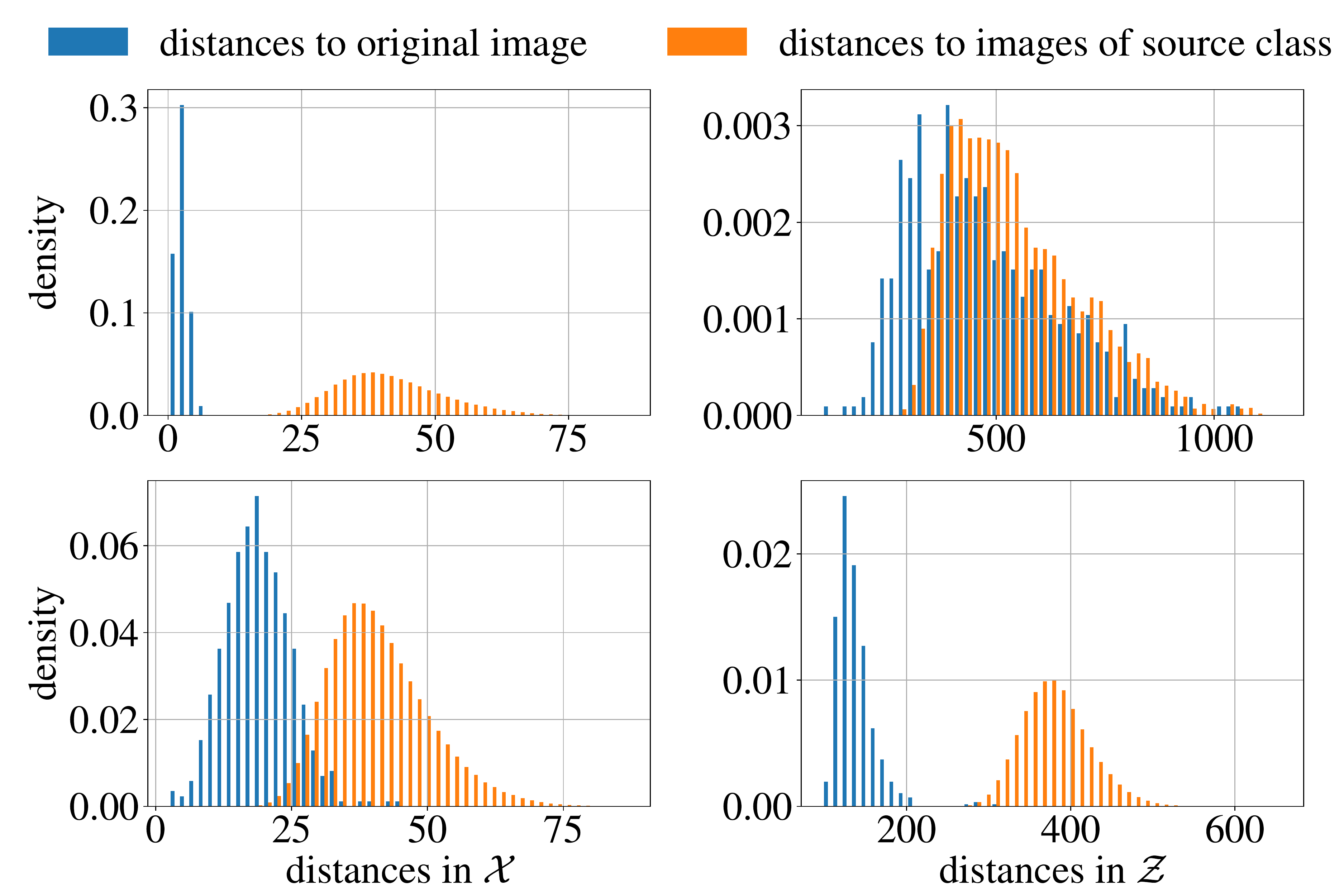}}
\caption{Euclidean distances in $\mathcal{X}$ and $\mathcal{Z}$ for adversarial examples (top row) and counterfactuals (bottom row) for the CelebA data set.}
\label{fig:L2_source_CelebA_conv_z}
\end{center}
\vskip -0.2in
\end{figure}

\begin{figure}[t]
\begin{center}
\centerline{\includegraphics[width=1.0\columnwidth]{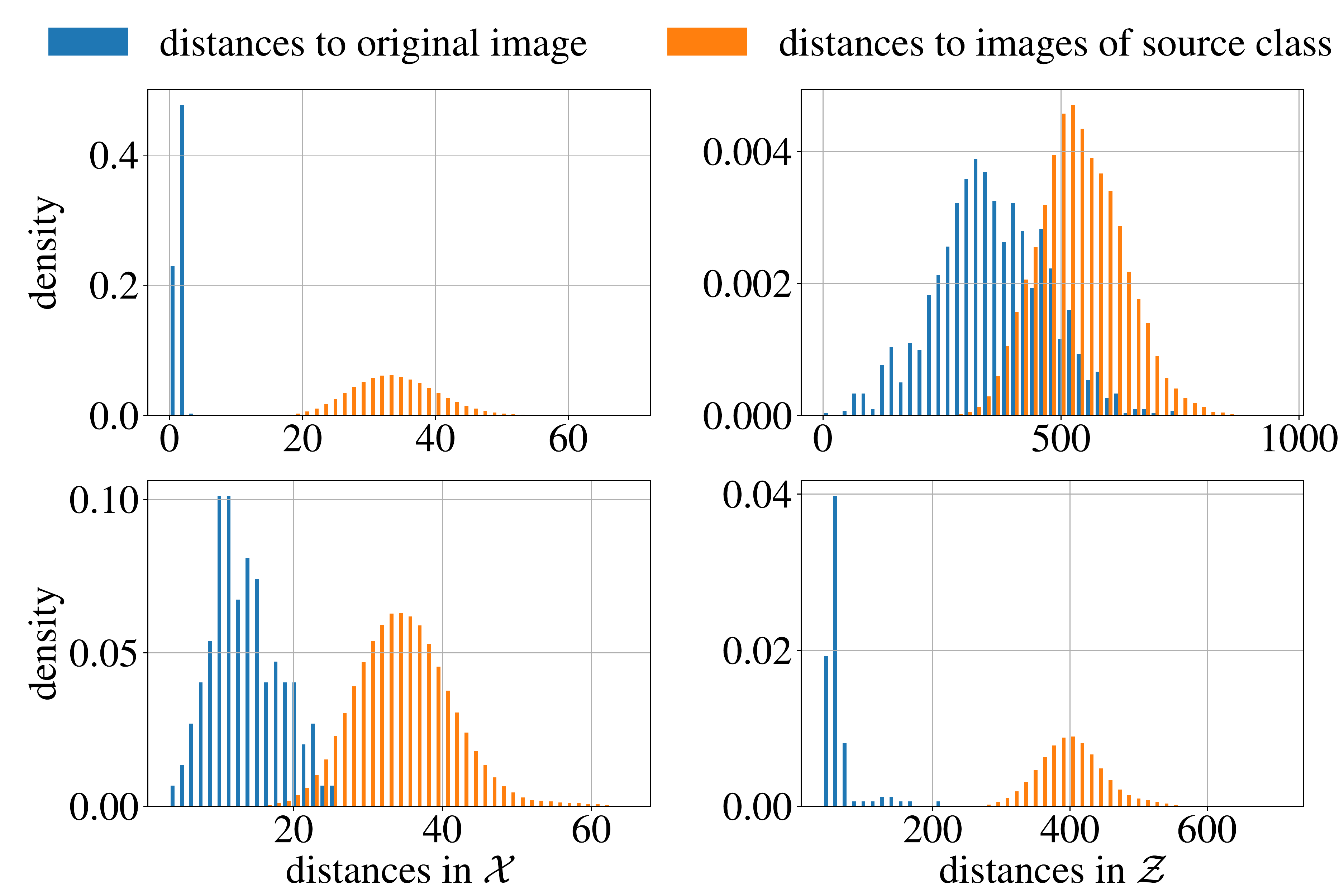}}
\caption{Euclidean distances in $\mathcal{X}$ and $\mathcal{Z}$ for adversarial examples (top row) and counterfactuals (bottom row) for the CheXpert data set.}
\label{fig:L2_source_CheXpert_conv_z}
\end{center}
\vskip -0.2in
\end{figure}

\begin{figure}[t]
\begin{center}
\centerline{\includegraphics[width=1.0\columnwidth]{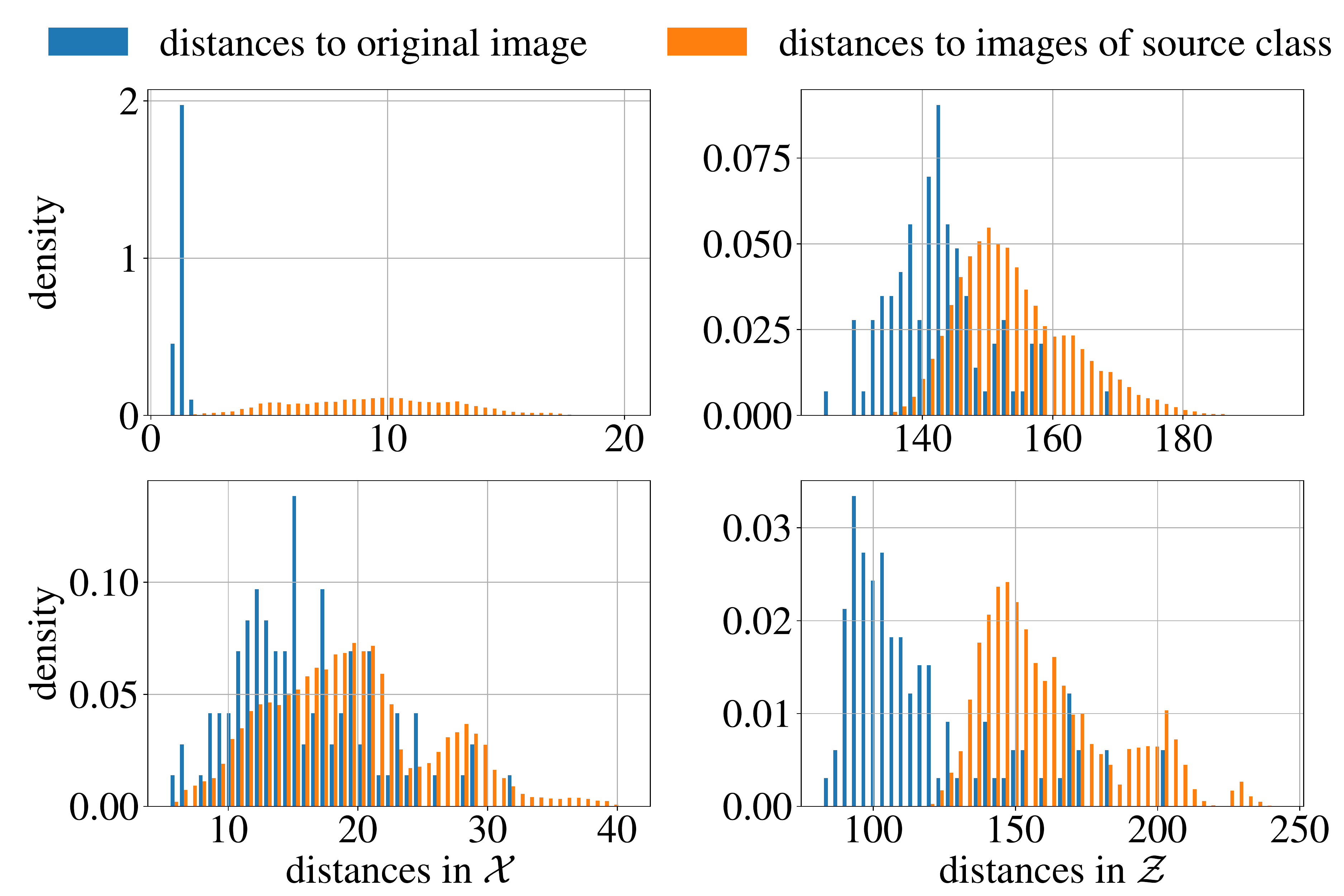}}
\caption{Euclidean distances in $\mathcal{X}$ and $\mathcal{Z}$ for adversarial examples (top row) and counterfactuals (bottom row) for the Mall data set. Source images have few pedestrians ($r<1$).}
\label{fig:L2_source_Mall_few_conv_z}
\end{center}
\vskip -0.2in
\end{figure}

\begin{figure}[t]
\begin{center}
\centerline{\includegraphics[width=1.0\columnwidth]{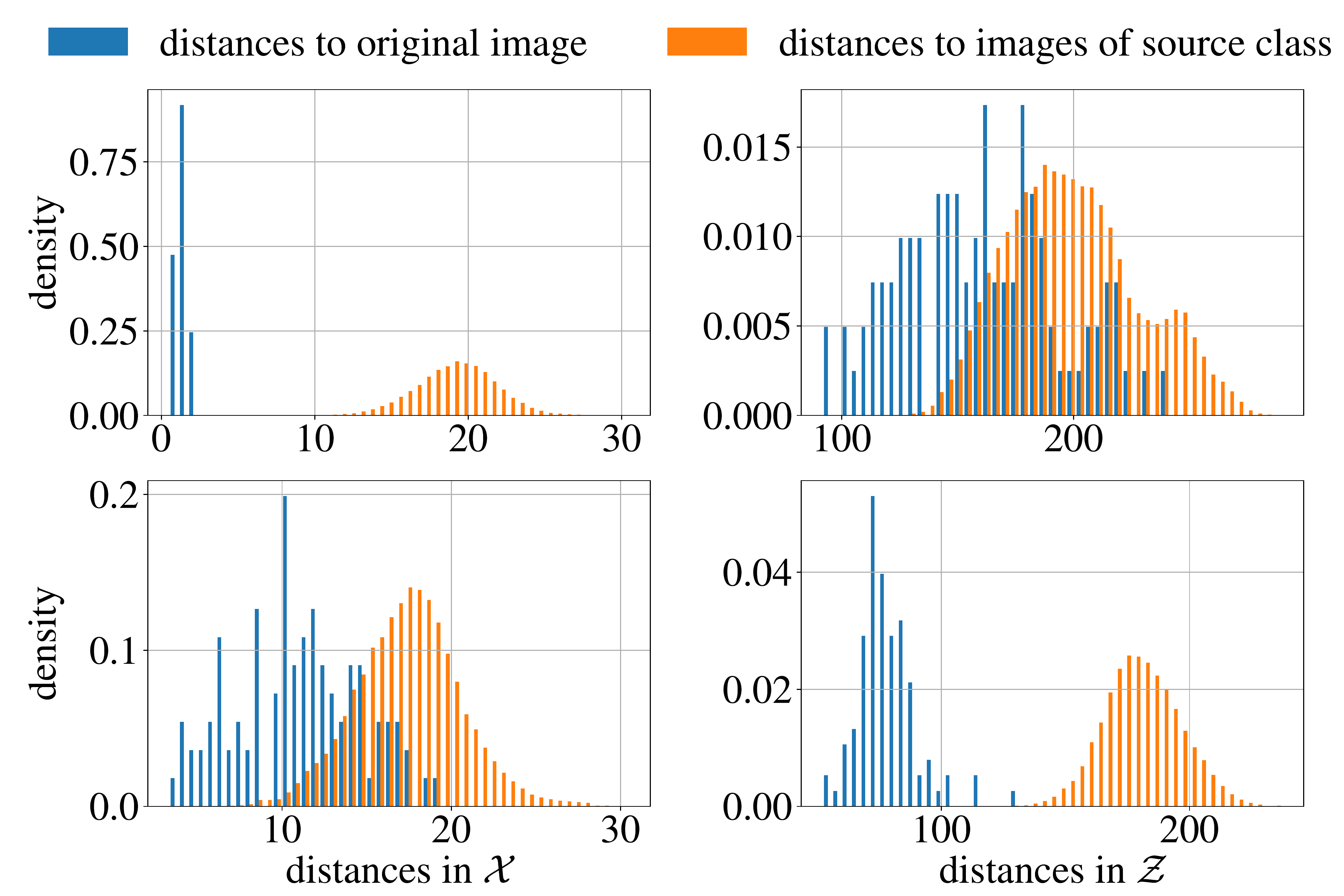}}
\caption{Euclidean distances in $\mathcal{X}$ and $\mathcal{Z}$ for adversarial examples (top row) and counterfactuals (bottom row) for the Mall data set. Source images have many pedestrians ($r>3$).}
\label{fig:L2_source_Mall_many_conv_z}
\end{center}
\vskip -0.2in
\end{figure}

To evaluate the proximity between counterfactuals/adversarials and their corresponding source images we calculate the Euclidean differences in $\mathcal{X}$ space as well as in $\mathcal{Z}$ space and compare them to the respective Euclidean differences for all images of the source class. For MNIST, CelebA and CheXpert we calculate the Euclidean differences for a maximum of 2000 test images for each counterfactual/adversarial. For the Mall data set we calculate the distances to 400 training images with $r<1$ when considering counterfactuals/adversarials for which we maximized $r$ and we calculate the distances to 400 training images with $r>3$ when considering counterfactuals/adversarials for which we minimized $r$. In addition we calculate all l2 differences between counterfactuals/adversarials and their respective source images. 
We show the distribution of distances to the respective source images and to all images of the source class in Figures~\ref{fig:L2_source_MNIST_conv_z},~\ref{fig:L2_source_CelebA_conv_z},~\ref{fig:L2_source_CheXpert_conv_z},~\ref{fig:L2_source_Mall_few_conv_z}, and~\ref{fig:L2_source_Mall_many_conv_z}.
As expected the Euclidean differences between adversarial examples and their respective source images are very small when measured in $\mathcal{X}$ space but a lot larger when measured in $\mathcal{Z}$ space. The Euclidean differences for counterfactuals measured in $\mathcal{X}$ are larger than those for adversarials but we can still observe that counterfactuals are significantly closer to their respective source image than to other images of the same class. The Euclidean distances measured in $\mathcal{Z}$ space are significantly smaller for counterfactuals than for adversarials, indicating that the latter may lie off manifold.
The effect is less pronounced for images from the Mall data set, as those have little variance in the background.

\subsection{Similarity between all images}

\begin{figure}[t]
\begin{center}
\centerline{\includegraphics[width=1.0\columnwidth]{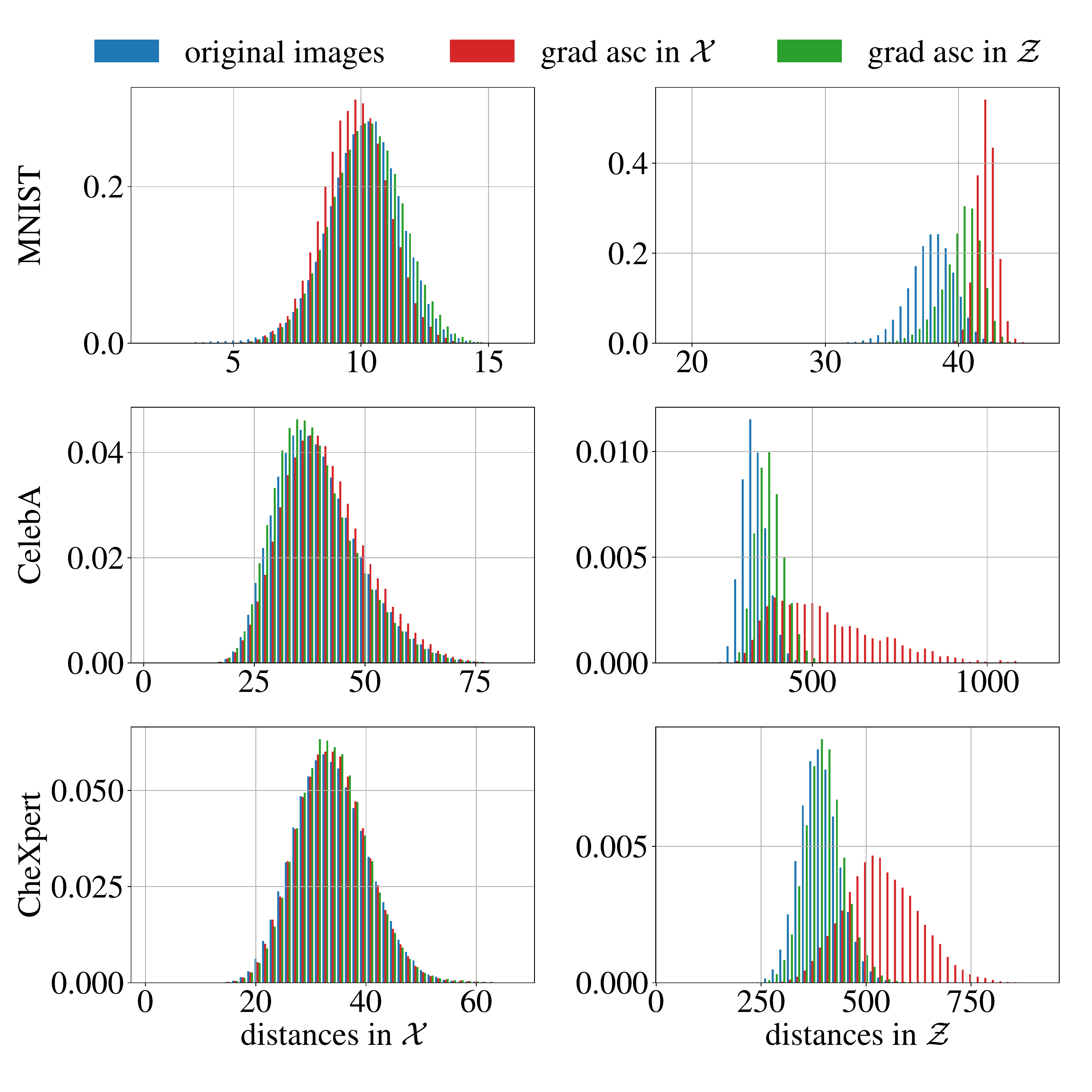}}
\caption{Distributions of Euclidean distances in $\mathcal{X}$ and $\mathcal{Z}$ for test images, adversarial examples and counterfactuals for three data set.}
\label{fig:L2_all_org_conv_z}
\end{center}
\vskip -0.2in
\end{figure}

We calculate the Euclidean distances in $\mathcal{X}$ and $\mathcal{Z}$ of randomly selected test images (all classes), adversarial examples and counterfactuals to randomly selected images from the training data set (all classes). Figure~\ref{fig:L2_all_org_conv_z} shows the distribution of distances for the data sets MNIST, CelebA and CheXpert. We note that for distances in $\mathcal{X}$ the distributions for original images, adversarial examples and counterfactuals are very similar while for distances in $\mathcal{Z}$ the distribution of distances for adversarial examples is notably shifted to the right, meaning that adversarial examples are further away from random data samples when the distances are measured in $\mathcal{Z}$, that is on the manifold. The effect is most notable for CelebA and CheXpert, for which the distances in $\mathcal{Z}$ of counterfactuals closely match the distribution of distances between images from the data set.

The original distribution of images from the Mall data set is strongly skewed towards few pedestrians. We can therefore not expect to achieve insights from comparing distributions of manipulated images.

\subsection{GANs}

\underline{MNIST:}
We use a Deep Convolutional GAN\footnote{\url{https://github.com/AKASHKADEL/dcgan-mnist}} that we train for 170 epochs using the Adam optimizer with weight decay of $5\times10^{-4}$ and learning rate of $2\times10^{-4}$. 

\underline{CelebA:} We use a progressively grown GAN\footnote{\url{https://github.com/rosinality/progressive-gan-pytorch}} and train on 600000 randomly selected training images.

\underline{CelebA-HQ:} We use the pretrained HyperStyle\footnote{\url{https://github.com/yuval-alaluf/hyperstyle}}~\cite{alaluf2021hyperstyle} network. 

\subsection{VAEs}
We use Adam without weight decay and with a learning rate of $\lambda=5\times 10^{-3}$ for all VAEs.

\underline{MNIST:}
We train a simple convolutional VAE for 80 epochs.
To evaluate the IM1 and IM2 measures we train two additional VAEs with the same structure on only the training images with label four and nine respectively for 100 epochs.

\underline{CelebA:} The VAEs for CelebA are only used to evaluate the IM1 and IM2 measures. We train 3 simple convolutional VAEs for 100 epochs on the complete training set, all training images with blond attribute equal to one and all training images with blond attribute equal to 0.

\subsection{Quantitative evaluation}
\underline{Nearest neighbours}: For MNIST, CelebA and CheXpert we search for the 10 nearest neighbours considering the complete test set (excluding the image the counterfactual/adversarial originated from), making sure there is an even distribution over classes present. For Mall, we search for the three nearest neighbours considering the complete training set (as the test set is relatively small and the data is unevenly distributed (many images with few pedestrians and very few with many pedestrians)).
We use the Euclidean norm as a distance measure.

\section{Examples for Counterfactuals}
In this appendix, we present results on randomly selected images from the four data sets for which we produce counterfactuals via the flow. For the heatmaps, we visualize both the sum over the absolute values of color channels as well as the sum over the color channnels.  
\begin{figure}[t]
\vskip 0.2in
\begin{center}
\centerline{\includegraphics[width=.8\columnwidth]{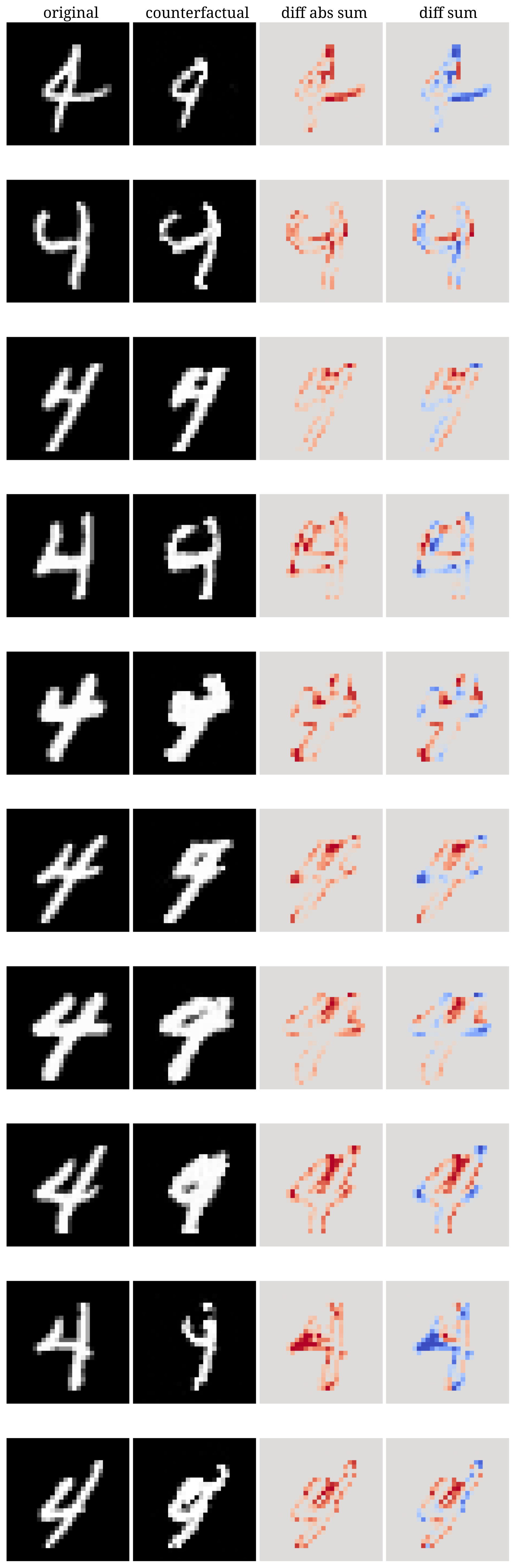}}
\caption{Randomly selected examples MNIST `four' to `nine'}
\label{fig:adv_attacks_MNIST}
\end{center}
\vskip -0.2in
\end{figure}

\begin{figure}[t]
\vskip 0.2in
\begin{center}
\centerline{\includegraphics[width=.8\columnwidth]{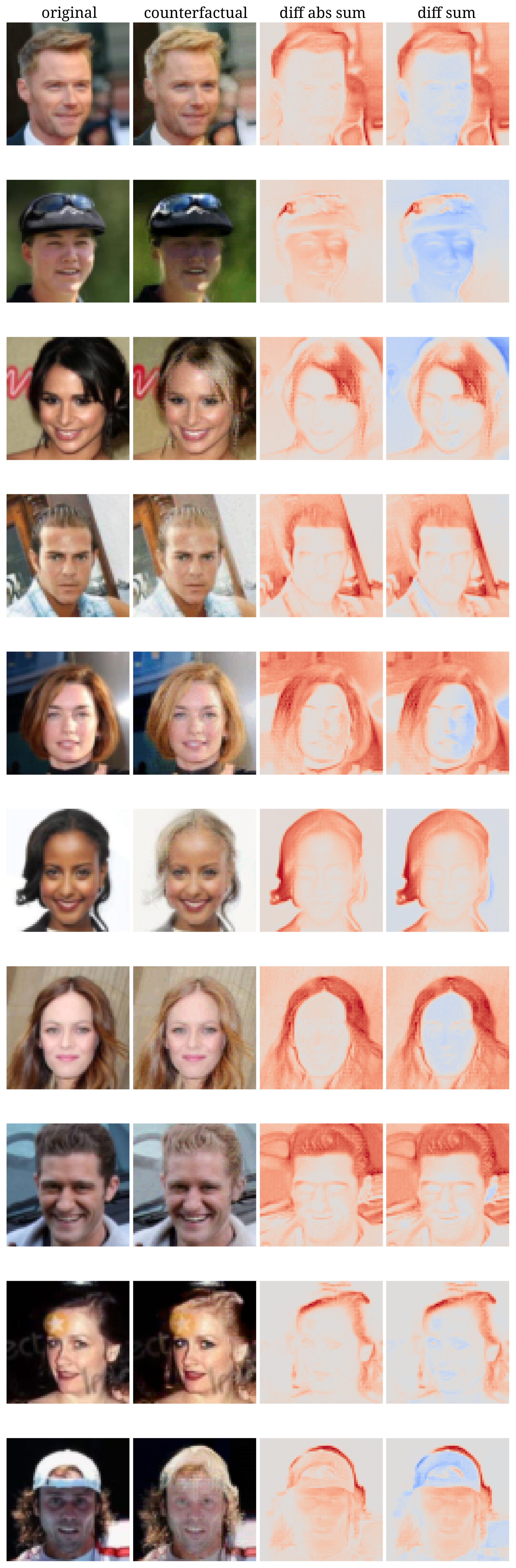}}
\caption{Randomly selected examples CelebA `not blond' to `blond'}
\label{fig:adv_attacks_CelebA}
\end{center}
\vskip -0.2in
\end{figure}

\begin{figure}[t]
\vskip 0.2in
\begin{center}
\centerline{\includegraphics[width=.8\columnwidth]{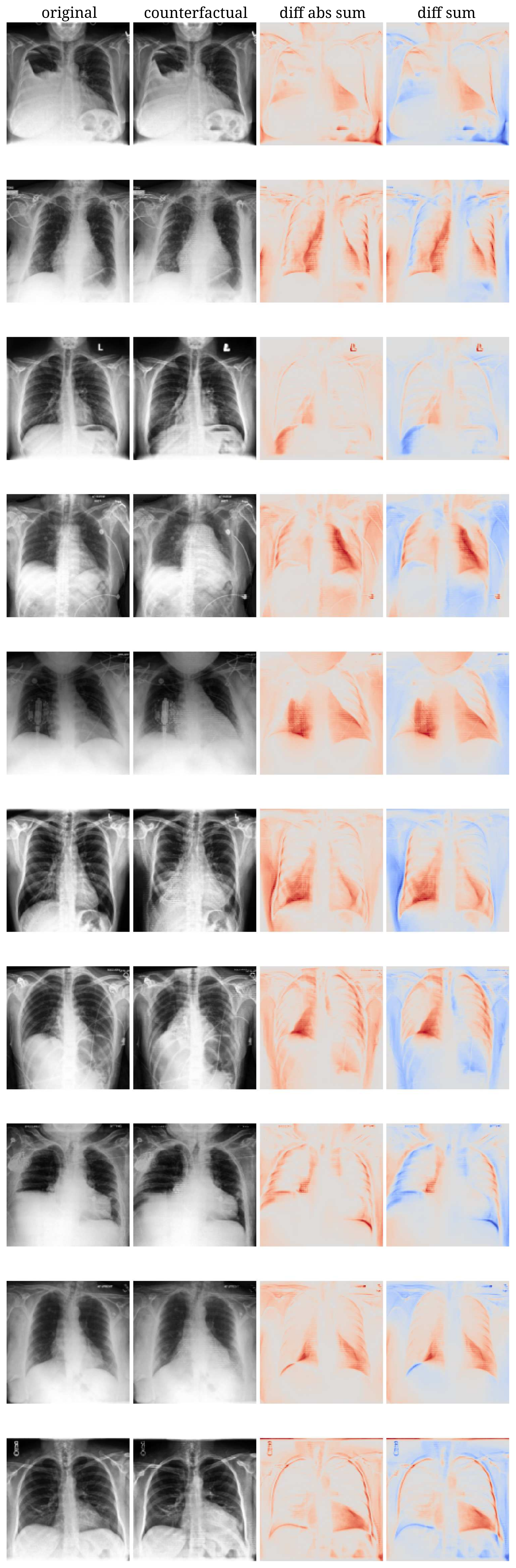}}
\caption{Randomly selected examples CheXpert `healthy' to `cardiomegaly'}
\label{fig:adv_attacks_CheXpert}
\end{center}
\vskip -0.2in
\end{figure}

\begin{figure}[t]
\vskip 0.2in
\begin{center}
\centerline{\includegraphics[width=.8\columnwidth]{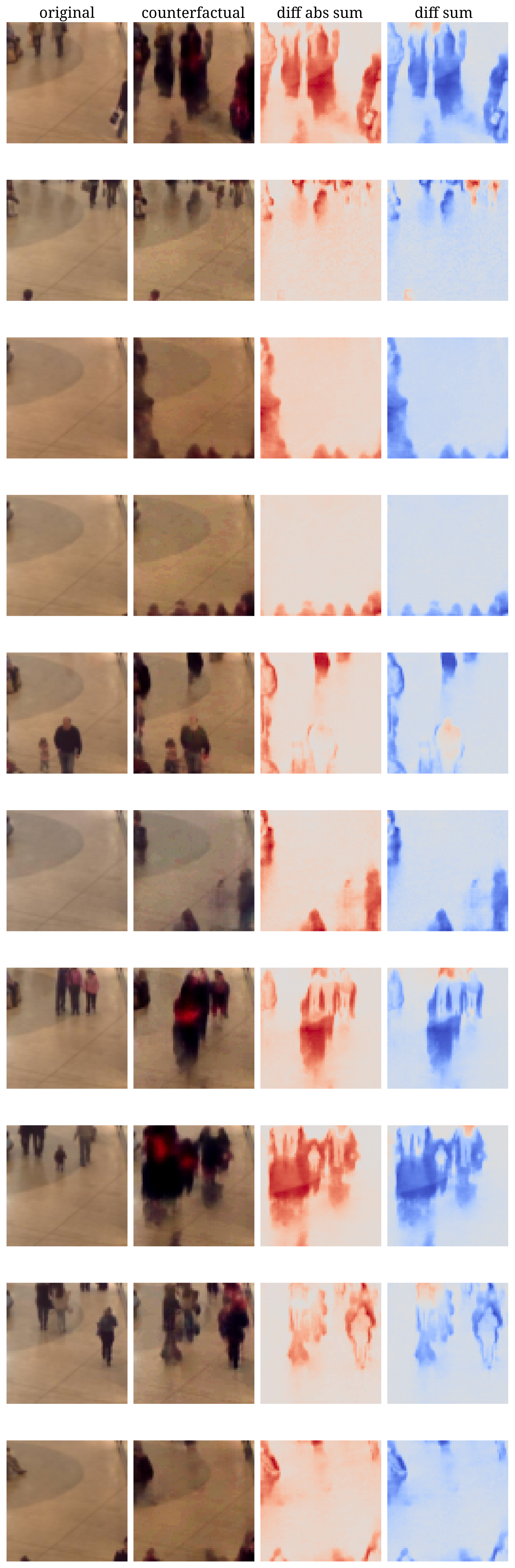}}
\caption{Randomly selected examples Mall `few people' to `10 people'}
\label{fig:adv_attacks_Mall_10}
\end{center}
\vskip -0.2in
\end{figure}

\begin{figure}[t]
\vskip 0.2in
\begin{center}
\centerline{\includegraphics[width=.8\columnwidth]{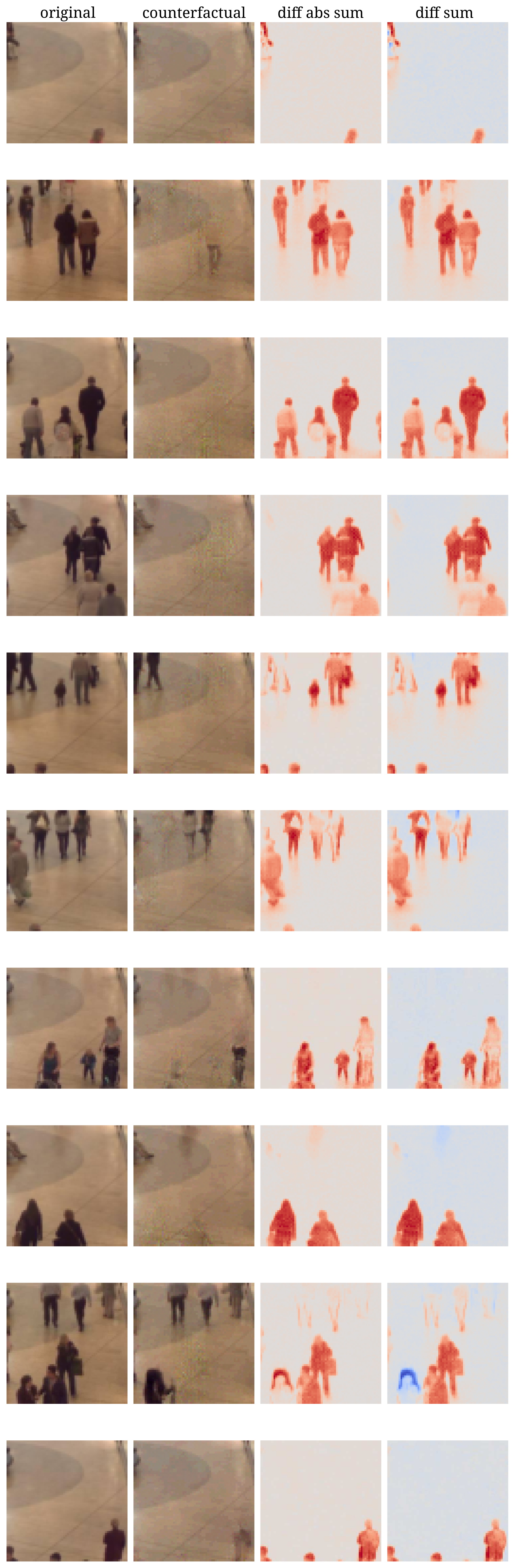}}
\caption{Randomly selected examples Mall `many people' to `0.01 people'}
\label{fig:adv_attacks_Mall_001}
\end{center}
\vskip -0.2in
\end{figure}


\end{document}